\documentclass{article}

\usepackage{arxiv}

\usepackage[utf8]{inputenc} % allow utf-8 input
\usepackage[T1]{fontenc}    % use 8-bit T1 fonts
\usepackage{hyperref}       % hyperlinks
\usepackage{url}            % simple URL typesetting
\usepackage{booktabs}       % professional-quality tables
\usepackage{amsfonts}       % blackboard math symbols
\usepackage{nicefrac}       % compact symbols for 1/2, etc.
\usepackage{microtype}      % microtypography
\usepackage{lipsum}		% Can be removed after putting your text content
\usepackage{graphicx}
\usepackage[numbers]{natbib}
\usepackage{doi}
\usepackage{amsmath}
\usepackage{cleveref}
\usepackage{todonotes}
\usepackage{subcaption}
\usepackage{lscape}

\title{Stacked networks improve physics-informed training: applications to neural networks and deep operator networks}

\date{} 					% Or removing it

\author{ 
        \href{https://orcid.org/0000-0002-6411-6198}{\includegraphics[scale=0.06]{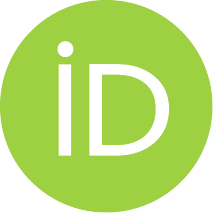}\hspace{1mm}Amanda A. Howard}\\
	Pacific Northwest National Laboratory\\
	Richland, WA 99354 \\
	\texttt{amanda.howard@pnnl.gov} \\
    \And
        \href{https://orcid.org/0009-0002-1337-0089}{\includegraphics[scale=0.06]{orcid.pdf}\hspace{1mm}Sarah H. Murphy} \\
        University of North Carolina\\
        Charlotte, NC 28223
    \And
        \href{https://orcid.org/0000-0001-5548-0265}{\includegraphics[scale=0.06]{orcid.pdf}\hspace{1mm}Shady E. Ahmed} \\
	Pacific Northwest National Laboratory\\
	Richland, WA 99354 \\
 	\texttt{shady.ahmed@pnnl.gov}  
    \And  
        \href{https://orcid.org/0000-0002-9928-5637}{\includegraphics[scale=0.06]{orcid.pdf}\hspace{1mm}Panos Stinis} \\
	Pacific Northwest National Laboratory\\
	Richland, WA 99354 \\
        \texttt{panagiotis.stinis@pnnl.gov}  
}

\renewcommand{\shorttitle}{Stacked networks improve physics-informed training}

\hypersetup{
}

\begin{document}
\maketitle

\begin{abstract}
Physics-informed neural networks and operator networks have shown promise for effectively solving equations modeling physical systems. However, these networks can be difficult or impossible to train accurately for some systems of equations. We present a novel multifidelity framework for stacking physics-informed neural networks and operator networks that facilitates training. We successively build a chain of networks, where the output at one step can act as a low-fidelity input for training the next step, gradually increasing the expressivity of the learned model.
The equations imposed at each step of the iterative process can be the same or different (akin to simulated annealing). The iterative (stacking) nature of the proposed method allows us to progressively learn features of a solution that are hard to learn directly. Through benchmark problems including a nonlinear pendulum, the wave equation, and the viscous Burgers equation, we show how stacking can be used to improve the accuracy and reduce the required size of physics-informed neural networks and operator networks. 
\end{abstract}

% keywords can be removed
\begin{figure}[h]
    \centering
    \includegraphics[width=\textwidth]{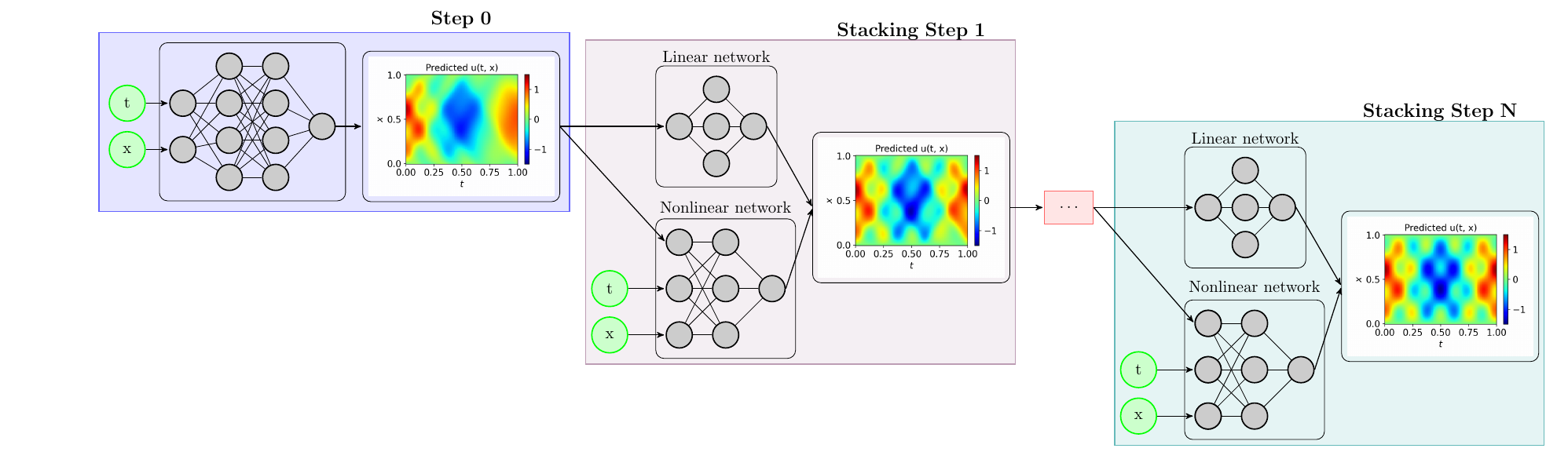}
    \caption{Graphical abstract. A previous prediction is used as the low-fidelity input for a physics-informed multifidelity neural network to generate a more accurate prediction as the output.  }
    \label{fig:Graphical_abstract}
\end{figure}

\section{Introduction} \label{sec:intro}
In recent years, a huge research focus has been on scientific machine learning methods for physical systems \cite{karniadakis2021physics, baker2019workshop, cuomo2022scientific}, for example, fluid mechanics and rheology \cite{jin2021nsfnets, raissi2020hidden, cai2021physics, Joglekar_2023, molina2023stokesian, brunton2021applying, dubois2022machine, shu2023physics}, metamaterial development \cite{liu2019multi, chen2020physics, fang2019deep}, high speed flows \cite{mao2020physics}, and power systems \cite{misyris2020physics, huang2022applications, moya2023dae, bento2023physics, wang2021predicting}, among many other applications. In particular, physics-informed neural networks, or PINNs \cite{raissi2019physics}, allow for accurately representing differential operators through automatic differentiation, leading to finding the solution to partial differential equations (PDEs) without explicit mesh generation. As physical systems often lack robust data, PINNs incorporate the differential equations of a system in the loss function of a neural network. Using automatic differentiation to minimize the loss function, the physics acts as a constraint for potential output solutions. The resulting models are able to train effectively using minimal training data.
% Recent work in scientific machine learning involves training physics-informed neural networks (PINNs) to approximate dynamical systems that model real-world phenomena. Although physical systems often lack robust data, PINNs incorporate the differential equations of a system in the loss function of a neural network. Using automatic differentiation to minimize the loss function, the physics act as a constraint for potential output solutions. The resulting models are able to train effectively using minimal training data \cite{raissi2019physics}. 
PINNs have been used successfully in a host of problems (see \cite{cuomo2022scientific} for a recent review and future perspectives).

In addition, operator learning techniques have recently received a great deal of focus due to their ability to represent maps between infinite-dimensional Banach spaces \cite{li2020fourier, wen2022u, li2020neural, lu2021learning}. Single fidelity deep operator networks (DeepONets) have shown success in a wide range of applications, see, for example, \cite{kumar2023real, goswami2022deep, yin2022interfacing, he2023hybrid, koric2023data, liu2022causality}. Physics-informed DeepONets can train to satisfy the solution of a differential equation, in a manner analogous to PINNs \cite{wang2021learning, goswami2022physicsB}, with successful applications including crack propagation \cite{goswami2022physics}, heat conduction \cite{koric2023data}, and instability-wave prediction \cite{hao2023instability}. 

% \SA{Suggestion--start introduction by referring to the lack of data in scientific applications, and hence the need for physics-informed training to mitigate this issue. Introduce PINN as a successful application, followed by operator learning and physics-informed operators. After that, we can list limitations of physics-informed training for PINNs and PI-DeepONets as below.}

Despite its promise, physics-informed training with little-to-no data can be challenging. For instance, PINNs tend to fail for dynamical systems for many possible reasons. The fixed points of a system, whether stable or unstable, create attractive optimal solutions that may not match the desired solution for a given initial condition \cite{rohrhofer2023ds}. Solutions to dynamical systems must fit the physical laws of the system, initial and boundary conditions, as well as data. Optimizing multiple objectives in this way can create issues, as minimizing the full loss function does not guarantee convergence with respect to each particular loss term. This can lead to a failure of the proposed solution to represent the solution specified by the initial or boundary conditions, which typically looks like a model training to the trivial solution. In a similar manner to PINNs, physics-informed DeepONets (PI-DeepONets) can be extremely difficult to train in the absence of data \cite{wang2023long}.

Since the introduction of PINNs (and later PI-DeepONets), several methods have been introduced to improve their training \cite{bajaj2021robust}. One portion of these efforts have been devoted to building unique architectures, including the works on multifidelity networks \cite{meng2019multifidelity} and finite basis domain decomposition schemes \cite{moseley2023finite, dolean2023multilevel}. Multifidelity PINNs \cite{meng2019multifidelity} traditionally use some data \cite{jagtap2022deep, chen2023feature}, in addition to a physics-informed term, to train more accurately than allowed by physics alone. In finite basis PINNs, neural networks approximate the solution to the differential equation as a finite set of basis functions with compact support \cite{moseley2023finite, dolean2023multilevel}. Similarly for DeepONets, when some low-resolution numerical data is available multifidelity DeepONets \cite{lu2022multifidelity, howard2023multifidelity, de2023bi} have been shown to improve the training for physics-informed problems.% \cite{howard2022onets}. 

%Along the lines of PINNs, several techniques have been proposed to improve the training of physics-informed DeepONets, including use of the neural tangent kernel \cite{wang2022improved}, and time marching schemes with transfer learning \cite{xu2023transfer}.

%These generally fall into several categories: unique architectures, intelligent choices of the points at which the terms in the loss function are evaluated, adaptive weighting schemes for the loss function, and causality and long-time integration schemes. In the first, recent papers have proposed multifidelity networks \cite{meng2019multifidelity} and finite basis domain decomposition schemes \cite{moseley2023finite, dolean2023multilevel}. Multifidelity PINNs \cite{meng2019multifidelity} traditionally use some data \cite{jagtap2022deep, chen2023feature}, in addition to a physics-informed term, to train more accurately than allowed by physics alone. In finite basis PINNs, neural networks approximate the solution to the differential equation as a finite set of basis functions with compact support \cite{moseley2023finite, dolean2023multilevel}. 

A second line of research has focused on the intelligent choices of the points at which the terms in the loss function are evaluated as well as adaptive weighting schemes for these terms.  For example, many methods have been proposed to adaptively sample the collocation points at which the PDE residuals are evaluated \cite{hou2023sampling,nabian2021efficient, wu2023comprehensive, gao2023failure, daw2022mitigating}. Moreover, it has been reported that the weighting schemes in the loss function can be very important for accurate training of PINNs. One key advancement in this direction is through self-adaptive weights \cite{mcclenny2020self}. The neural tangent kernel has also been used with great success both for PINNs \cite{wang2022and} and PI-DeepONets \cite{wang2022improved}. Finally, several works have looked at using long-time integration, time-stepping, or causality schemes, where the training is performed in several successive steps, possibly with transfer learning in place \cite{wang2022causality, wight2020solving,mattey2022novel, mojgani2022lagrangian, penwarden2023unified,xu2023transfer}.

While these methods show great promise, there are still cases where PINNs and PI-DeepONets can fail to train, as noted in many recent works \cite{rohrhofer2023ds, chuang2023predictive}. In this work, we present a simple, yet very effective, \emph{multifidelity stacking} approach for learning dynamical systems. We iteratively train a multifidelity PINN/PI-DeepONet for a user-defined number of steps, where the low-fidelity model at each step takes the output of the previous step as input (see Fig. \ref{fig:Graphical_abstract}). Although we adopt the notion of multifidelity networks, we highlight that we do not use any data in our approach (except for the given initial and boundary conditions). Instead, the multifidelity architecture aims at building a chain of networks, where each link represents the lower fidelity model for the next one. The iterative training aims to progressively refine the predictive ability of a PINN/PI-DeepONet. We also note that during the iterative training we can enforce the same equation for all iterative steps or different equations. This allows to begin the training from a simpler problem and gradually morph it, through iterations, to the original (and harder) problem, similar to simulated annealing \cite{bertsimas1993simulated} and curriculum learning \cite{krishnapriyan2021characterizing} approaches. Finally, we want to draw attention to a similar but  alternative approach, called Galerkin Neural Networks (GNN), for iteratively building neural network approximations of solutions for variational formulations of partial differential equations \cite{ainsworth2021galerkin, ainsworth2022galerkin}. The main difference between Galerkin Neural Networks and the current work is in the criterion used to select either a correction or the full solution at each iteration. 

The paper is organized as follows. We first introduce the stacking method in Section \ref{sec:method}. We then discuss illustrative examples in Section \ref{sec:PINN} that show how the stacking method can be used for multiscale problems to train in cases where it is not possible to train a standard PINN, and also to train accurately with fewer trainable parameters than needed for a standard PINN. In Section \ref{sec:PINN} we also consider the wave equation, which shows how stacking PINNs can be used to accurately morph from an equation that is easy to train to an equation that is difficult to train. In Section \ref{sec:deeponet} we show the extension of stacking PINNs to stacking PI-DeepONets. Finally, we summarize key takeaways from the current work as well as possible directions for future research in Section~\ref{sec:conclusion}.

\section{Method}\label{sec:method}
While the principle behind the stacking approach is the same for PINNs and PI-DeepONets, we choose to present in this section the details mostly for PINNs, delegating some details about the application to PI-DeepONets to Subsection~\ref{sec:method_deeponet}.

\subsection{Physics-informed neural networks}
PINNs \cite{raissi2019physics} allow for finding solutions to physical systems and discovery of ordinary or partial differential equations with limited training data. PINNs incorporate the dynamics of the system in the loss function of the network to enforce physical laws or domain expertise. Specifically, we define a system over an open, bounded domain $\Omega \in \mathbb{R}^n$ with boundary $\partial\Omega$ as follows:
\begin{align}
    &s_t+\mathcal{N}_x[s]=\textbf{0}, & x\in \Omega, t\in [0,T], \\
    &s(x,t)=g(x,t), & x\in \partial\Omega, t\in [0,T], \\
    &s(x,0)=u(x), & x\in \Omega, 
\end{align}
where $g$ and $u$ are given functions for the boundary and initial conditions, respectively while $x$ denotes the spatial coordinates, $t$ is the temporal coordinate, and $\mathcal{N}_x$ is a differential operator with respect to $x$. Our goal is to utilize deep neural networks (DNNs) to approximate $s(x,t)$. The parameters (weights and biases) for the network are denoted by $\theta$, which are optimized to minimize the following loss function:
\begin{equation}
\mathcal{L}(\theta)=\lambda_{ic}\mathcal{L}_{ic}(\theta)+\lambda_{bc}\mathcal{L}_{bc}(\theta)+\lambda_r\mathcal{L}_r(\theta), \label{eq:loss}
\end{equation}
with subscripts $ic, bc, r$ corresponding to loss terms with respect to initial conditions, boundary conditions, and residual. We sample $N_{ic}$, $N_{bc}$, and $N_r$ points in the space-time domain for the initial conditions, boundary conditions, and residual collocation points, respectively. The training data sets are denoted by $\{(x_{ic}^i), u(x_{ic}^i)\}_{i=1}^{N_{ic}}, \{(x_{bc}^i, t_{bc}^i), g(x_{bc}^i, t_{bc}^i)\}_{i=1}^{N_{bc}},$ and $ \{(x_{r}^i, t_{r}^i)\}_{i=1}^{N_{r}}.$ A common choice for the terms in the loss function is through the mean squared errors (MSEs) according to
\begin{align}
    \mathcal{L}_{ic}(\theta) &= \frac{1}{N_{ic}} \sum_{i = 1}^{N_{ic}} \left[ \mathcal{F}(x_{ic}^i, 0; \theta) - u(x_{ic}^i) \right]^2, \\
    \mathcal{L}_{bc}(\theta) &= \frac{1}{N_{bc}} \sum_{i = 1}^{N_{bc}} \left[ \mathcal{F}( x_{bc}^i, t_{bc}^i; \theta) - g(x_{bc}^i, t_{bc}^i)\right]^2, \\
    \mathcal{L}_{r}(\theta) &= \frac{1}{N_{r}} \sum_{i = 1}^{N_{r}} \left[ \mathcal{F}_t(x_{r}^i, t_{r}^i; \theta)+ \mathcal{N}_x[\mathcal{F}( x_{r}^i, t_{r}^i; \theta)]\right]^2,   
\end{align}
where $\mathcal{F}(x, t; \theta)$ denotes the output of the network with parameters $\theta$ at $(x, t)$.

As noted in \ref{sec:intro}, the selection of the weights $\lambda_{ic}$, $\lambda_{bc}$, and $\lambda_{r}$ can have an impact on physics-informed training and several methods have been developed for adaptively choosing these weights, including soft-attention mechanism weights \cite{mcclenny2020self}, the neural tangent kernel weights \cite{wang2022and}, among other variations \cite{qadeer2023efficient, hou2023enhancing, xiang2022self}. However, we have opted to hand-pick these terms based on knowledge of the system being learned instead of focusing on the details of adaptive weight selection, which is not the focus of this paper. A typical rule-of-thumb that we follow for selecting these weights is $\lambda_{ic} = \lambda_{bc}=10\lambda_{r}$, with the exact values used in the examples shown here given in Appendix \ref{sec:appendix_trainingparams}. Nonetheless, we note that any of the aforementioned adaptive weighting methods could be applied in training stacking PINNs seamlessly.%While we do not fully explore the use of these adaptive methods in this work, we note that they could be applied in training stacking PINNs seamlessly. 

\begin{figure}[h]
    \centering
    \includegraphics[width=.6\textwidth]{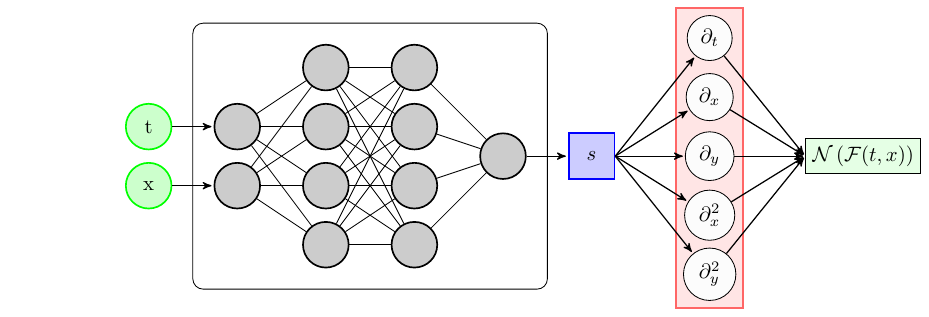}
    \caption{Physics-informed neural network}
    \label{fig:PINN}
\end{figure}

\subsubsection{When PINNs fail to train}\label{sec:pinn_fails}
Despite the promising results of PINNs, there exist some important examples for which PINNs fail to train with the standard framework, such as a pendulum with damping and the wave equation \cite{rohrhofer2023ds, chuang2022experience, chuang2023predictive, krishnapriyan2021characterizing}. For dynamical systems in particular, solutions must fit the physical laws on the system, initial and boundary conditions, as well as any given data. With complex loss functions, minimizing the full function does not ensure convergence for each particular loss term. 

\begin{figure}[h]
    \centering
    \includegraphics[width=\textwidth]{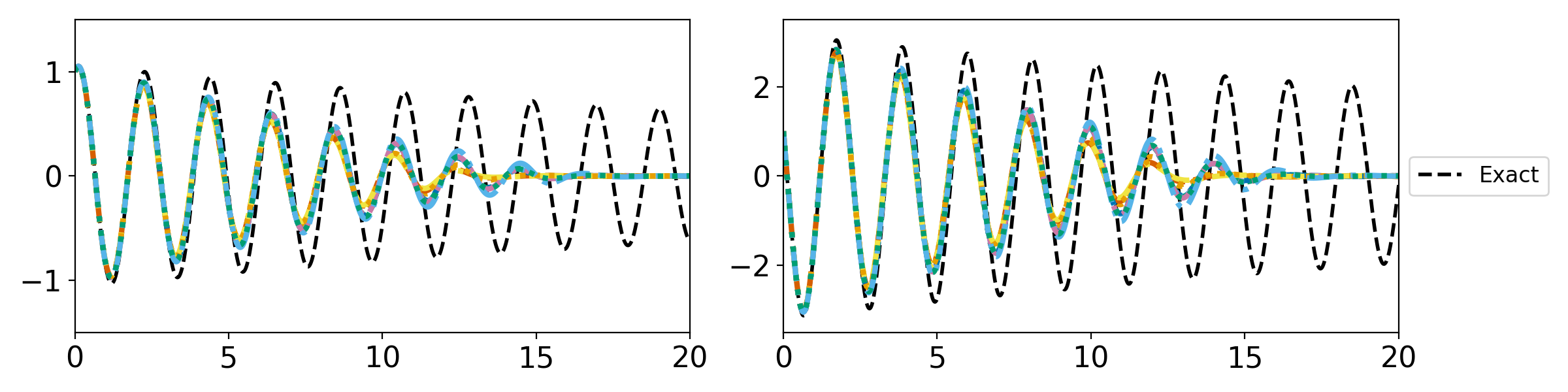}
    \caption{PINN results $s_1$ (left) and $s_2$ (right) as a function of time for the pendulum problem for ten random initial seedings. In each case, the PINN solution decays and does not agree well with the exact solution. %\SA{The labels for the x and y axes are missing.}
    }
    \label{fig:SF_pen}
\end{figure}

A simple example, yet one where a PINN struggles significantly to train, is a damped pendulum, shown in Fig. \ref{fig:SF_pen}. The system is governed by a system of two first-order ordinary differential equations (ODEs) for $t \in [0, T]$ 
\begin{align}
    \frac{d s_1}{dt} &= s_2, \label{eq:pendulum_1}\\
    \frac{d s_2}{dt} &= -\frac{b}{m} s_2 - \frac{g}{L} \sin(s_1), \label{eq:pendulum_2}
\end{align}
where $s_1$ and $s_2$ are the position and velocity of the pendulum, respectively. The initial conditions are chosen as $s_1(0) = s_2(0) = 1$. We take $m=L=1$, $b=0.05$, $g=9.81$, and $T=20$ as in \cite{wang2023long}.

While the trained outputs from the ten PINNs shown in Fig. \ref{fig:SF_pen} satisfy the initial condition, the solutions decay to zero past $t \approx 14$. We note that $s_1 = s_2 = 0$ would be a true solution for the pendulum if the initial conditions were $s_1(0) = s_2(0) = 0$. Thus, once the solution decays to zero, the loss term associated with the residual is minimized, however, the solution is not correct. 
 
\subsection{Multifidelity PINNs}
One method used to improve the training of PINNs is multifidelity learning, where physics-informed cases can be augmented by a small amount of data or additional physics knowledge to increase the accuracy of the final solution. Many approaches and architectures for multifidelity PINNs have been proposed for a wide range of applications, e.g., \cite{meng2019multifidelity, liu2023multi, penwarden2022multifidelity, aliakbari2022predicting, ramezankhani2022data, regazzoni2021physics}. In this section we summarize the approach taken in this paper, based on the approach in \cite{meng2019multifidelity}. 

 %As the stacking method involves iteratively applying multifidelity PINNs, the low fidelity approximation at step 0 is given first by an initial single fidelity PINN, and then at step $i$ by the output of the multifidelity network at step $i-1.$
 
\begin{figure}[h]
    \centering
    \includegraphics[width=0.7\textwidth]{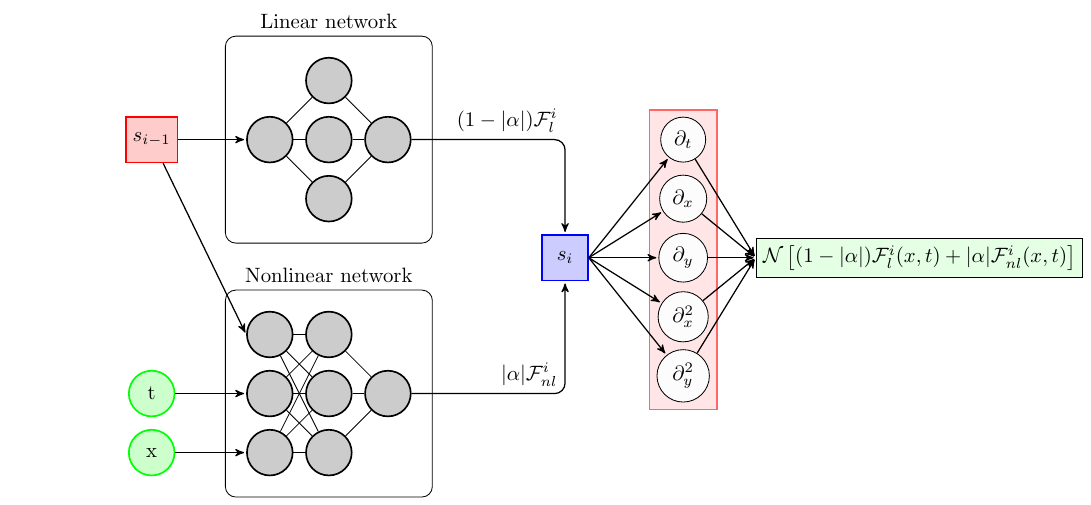}
    \caption{Multifidelity physics-informed neural network}
    \label{fig:MF_PINN}
\end{figure}

A multifidelity PINN consists of two neural networks, which are trained simultaneously to learn the nonlinear, $\mathcal{F}_{nl},$ and linear, $\mathcal{F}_l,$ correlations between the low-fidelity approximation and the high-fidelity approximation of the system, as shown in Fig. \ref{fig:MF_PINN}. The low-fidelity approximation can be the prediction of a neural network trained with low-fidelity data as in \cite{meng2019multifidelity}. The output of the multifidelity PINN is the convex combination of the learned correlation networks
\begin{equation}
    \mathcal{F}_{MF}(x,t; \theta)=|\alpha| \mathcal{F}_{nl}(x,t;\theta)+(1-|\alpha|)\mathcal{F}_l(x,t,\theta),
\end{equation}
where the set of trainable parameters of each network is denoted by $\theta$ and $\alpha$ is a trainable coefficient that controls the interplay between the linear and nonlinear components.  To enforce learning the linear correlation, $\mathcal{F}_l$ does not use an activation function. The multifidelity loss function is given by: 
\begin{equation}
\mathcal{L}_{MF}(\theta)=\lambda_{ic}\mathcal{L}_{ic}(\theta)+\lambda_{bc}\mathcal{L}_{bc}(\theta)+\lambda_r\mathcal{L}_r(\theta) + \alpha^4. \label{eq:loss_MF}
\end{equation}
The penalty term $\alpha^4$ is chosen so that $\alpha$ is small, and the network predominately learns a linear correlation when possible. By maximizing the linear correlatin, the nonlinear network can be smaller, reducing the number of trainable parameters. 

Previous work with multifidelity DNNs and multifidelity PINNs \citep{howard2023continual, meng2019multifidelity} have chosen to add a regularization term that penalizes the sum of the squares of the weights and biases of the nonlinear correlation network, instead of the penalty term in Eq. \ref{eq:loss_MF}. The goal of such a term was to prevent overfitting of the nonlinear correlation, as well as enforce learning a linear correlation when possible. In our tests, this regularization did not perform as well, perhaps because we train PINNs without data, so overfitting to data is not a concern. In addition, whether we use the exact same equation and loss function at every stacking level or we allow small modifications between levels, we expect that the linear correlation should be quite strong, in contrast with data-driven multifidelity training which uses low- and high-fidelity datasets for the low- and high-fidelity models.

As a very general concept, multifidelity PINNs are flexible and can use data or physics-informed training for both the low- and high-fidelity approximations. For example, data generated by a low order numerical model could be used as low-fidelity data because it will have a high amount of error. Then, physics can be enforced for the high-fidelity approximation using physics-informed training. In another case, a low order approximation of the physics could be enforced for the low-fidelity training, and a high order approximation of the physics could be used for the high-fidelity training. The advantage of multifidelity PINNs is that they allow for combinations of the available knowledge about the problem, producing more accurate and robust approximations than training with the low-fidelity or high-fidelity data or physics alone. We leverage this flexibility and consider the possibility of the low-fidelity component being a multifidelity PINN on its own, and gradually build up a stack of multi- multifidelity PINNs as described in \ref{sec:method_stacking}.

\subsection{Stacking PINNs} \label{sec:method_stacking}
\begin{figure}[h]
    \centering
    \includegraphics[width=.9\textwidth]{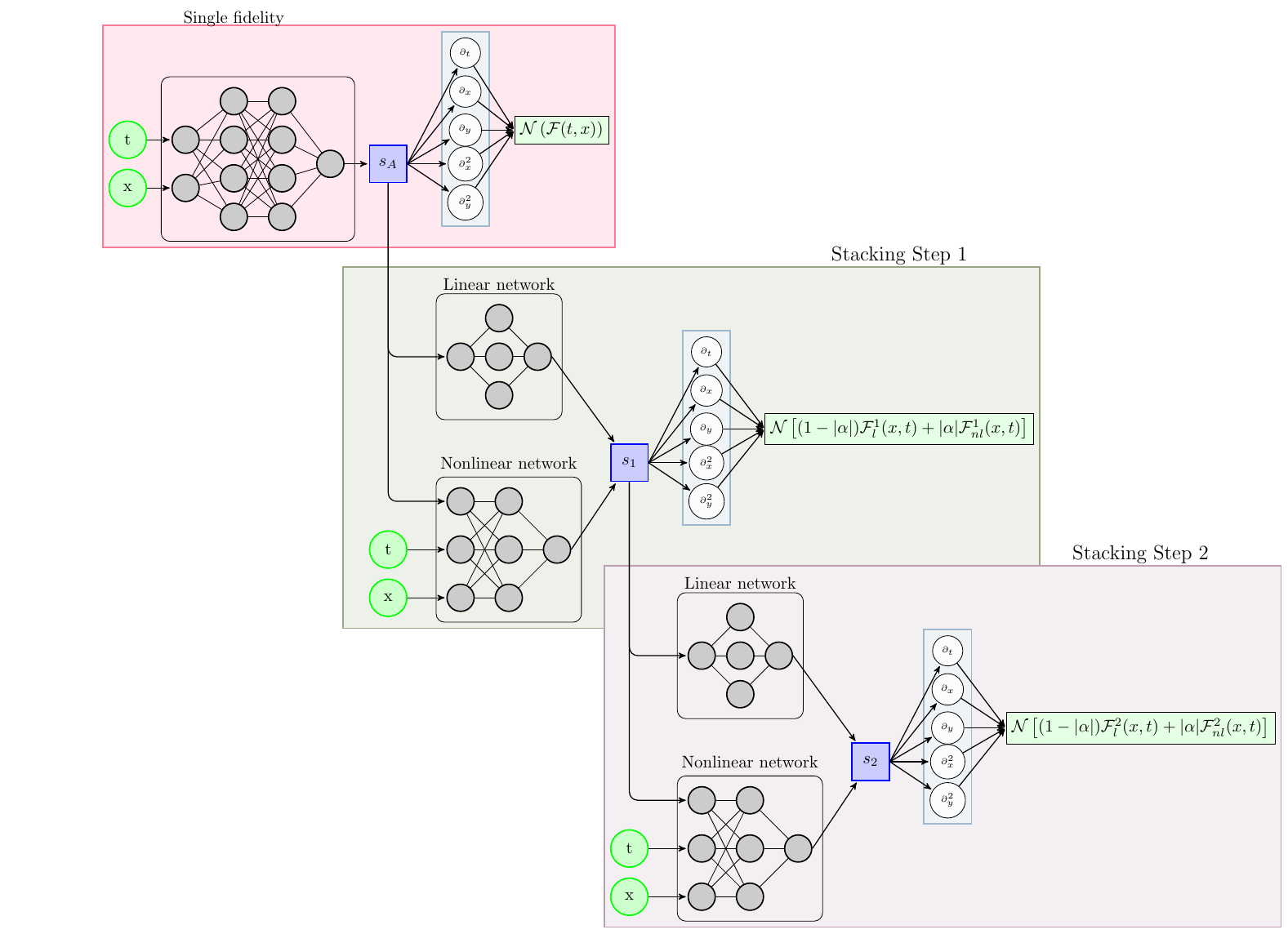}
    \caption{Stacking multifidelity physics-informed neural network}
    \label{fig:stacking_PINN}
\end{figure}

We propose the use of \emph{stacking PINNs}, in which multiple multifidelity PINNs are ``stacked'', so that the output of each multifidelity PINN is taken as the low-fidelity approximation for a new multifidelity PINN, as illustrated in Fig. \ref{fig:stacking_PINN}. As we show in this paper, stacking PINNs allow for more expressive solutions and can train for cases where single fidelity PINNs and traditional multifidelity PINNs cannot reach a satisfactory solution. In addition, stacking PINNs can use smaller network sizes, reducing the total number of trainable parameters needed to reach a given error. 

The process is implemented through the following steps:
\begin{enumerate}
    \item Step 0: Train a single fidelity PINN  to approximate the solution. This network is denoted $\mathcal{F}^0(x,t;\theta^0).$
    \item Stacking steps: for step $i>0$, train a multifidelity PINN $\mathcal{F}^i(x, t; \theta^i)$ that takes the output from the previous step $\mathcal{F}^{i-1}(x,t;\theta^{i-1})$ as a low fidelity approximation.
\end{enumerate}

One advantage of the stacking method is that we can use information from previously trained levels as a prior to inform the training of the current level. For example, we transfer the weights from previously trained networks to the current network instead of initializing the current network parameters randomly, akin to transfer learning. %In this work, each stacking network is initialized using the previously trained parameters to minimize the training time. 
In notation $\theta^i_0 = \theta^{i-1}$ and $\alpha^i_0 = \alpha^{i-1}$ for each stacking step $i > 1$, where $\theta^i_0$ denotes the initial values of the trainable parameters. We also note that, unlike regular multifidelity PINNs where the low- and high-fidelity networks are often trained simultaneously, we perform the stacking process sequentially. In other words, once a network is trained at level $i$, its parameters ($\theta^i$ and $\alpha^i$) are frozen throughout the next stacking steps. This alleviates the computational burden of optimizing a large number of parameters at once. Furthermore, it allows us to progressively increase the number of stacking levels, e.g., using sanity checks to determine whether more refinement is needed, without the need to retrain the whole stack.

\subsection{Deep operator networks} \label{sec:method_deeponet}
Physics-informed operator training allows for a great deal more expressiveness than PINNs because the solution is learned as the map between two Banach spaces. While PINNs learn the solution for a single initial condition and boundary condition, PI-DeepONets learn the solution for a family of initial conditions forming a Banach space. Several methods of operator learning are currently extremely popular \cite{li2020fourier, lu2021learning, li2020neural, wen2022u}. In this paper, we will focus on DeepONets \cite{lu2021learning}. 

Consider a general parametric PDE of the form 
\begin{align}
    \mathcal{N}(u, s) &= 0
\end{align}
with boundary conditions 
\begin{equation}
    \mathcal{B}(u, s) = 0.
    \end{equation}
The PDE solution is denoted by $\mathcal{G}(u) = s(u)$ for $\mathcal{G} \; : \mathcal{U} \rightarrow \mathcal{S}$, where $\mathcal{U}$ is the space of input parameters, generally the initial conditions, with $u \in \mathcal{U}$, and $\mathcal{S}$ is the space of PDE solutions on a domain $\Omega$. $s \in \mathcal{S}$ is an unknown function governed by the PDE system. 

A standard DeepONet consists of two DNNs, the branch and the trunk. The branch and trunk networks are trained simultaneously and combined in a dot product to express the solution. The input to the branch network is the function $u\in \mathcal{U}$ discretized at a discrete set of $M$ points, which are at fixed locations for each sample in the training and test set. The input to the trunk network is the independent variables, typically time and spatial coordinates. The output of the DeepONet is denoted by
\begin{equation}
    \mathcal{G}_\theta(\mathbf{u})(\mathbf{x}) = \sum_{k=1}^p b_k(u_1, \ldots, u_M)t_k(\mathbf{x})
\end{equation}
where $\theta$ denotes the trainable parameters of the unstacked DeepONet \cite{lu2019deeponet, lu2021learning}. 

As in our previous work \cite{howard2023multifidelity}, we choose to use ``modified'' DeepONets \cite{wang2022improved}, which introduce encoder layers for the branch and trunk nets. At each hidden layer, the branch and trunk are combined in a convex combination with the encoder layers. This modification has been shown to increase the accuracy of DeepONet training \cite{wang2022improved}. 

Physics-informed DeepONets (PI-DeepONets) are trained in much the same manner as PINNs, with similar loss functions. The loss function is given by
\begin{equation}
    \mathcal{L}_{DoN}(\theta) = \lambda_{bc} \mathcal{L}_{bc}(\theta) + \lambda_{r} \mathcal{L}_{r}(\theta) \label{eq:DeepONet_loss}
\end{equation}
where $\lambda_{bc}$ and $\lambda_{r}$ are tunable weighting terms and $\mathcal{L}_{bc}(\theta)$ and $\mathcal{L}_{res}(\theta)$ are the MSEs in satisfying the PDE system, given by
\begin{align}
    \mathcal{L}_{r}(\theta) &= \frac{1}{N}\frac{1}{R}\sum_{i=1}^N\sum_{j=1}^R \left[ \mathcal{N}\left(u^i, \mathcal{G}_\theta(\mathbf{u}^i)(x_j^i)\right) \right]^2\\
        \mathcal{L}_{bc}(\theta) &= \frac{1}{N}\frac{1}{R_{bc}}\sum_{i=1}^N\sum_{j=1}^{R_{bc}} \left[ \mathcal{B}\left(u^i, \mathcal{G}_\theta(\mathbf{u}^i)(x_j^i)\right) \right]^2
\end{align}
The number of collocation points are $R$ for the residual term and $R_{bc}$ for the boundary condition terms, which are randomly sampled in the domain $\Omega$ and the domain of $\mathcal{B}$, and denoted by $\{x_j^i\}_{i=1}^r$ and $\{x_j^i\}_{i=1}^{R_{bc}}$. 

We follow the multifidelity DeepONet framework introduced in \cite{howard2023multifidelity}, where the correlation is learned between the low-fidelity output and a linear DeepONet and nonlinear modified DeepONet. The output is given by the sum of the two networks, 
\begin{equation}
    \mathcal{G}_\theta^{MF}(\mathbf{u})(\mathbf{x}) = 
       |\alpha| \mathcal{G}_\theta^{nl}(\mathbf{u})(\mathbf{x}) + 
       (1-|\alpha|) \mathcal{G}_\theta^{l}(\mathbf{u})(\mathbf{x}), 
\end{equation}
and the loss function is modified to 
\begin{equation}
    \mathcal{L}_{MF, DoN}(\theta) = \lambda_{bc} \mathcal{L}_{bc}(\theta) + \lambda_{r} \mathcal{L}_{r} + \alpha^4. \label{eq:MF_DeepONet_loss}
\end{equation}

 \subsubsection{Neural tangent kernel}
Choosing the weighting terms ($\lambda_i$'s) in the loss functions Eqs. \ref{eq:loss_MF} and \ref{eq:DeepONet_loss} can be difficult, and can require expensive hand tuning for accurate training. The neural tangent kernel (NTK) \cite{allen2019convergence, jacot2018neural,du2019gradient} is a recently proposed method for determining the optimal weights, with the advantage that the weights are adaptive in space. The NTK has been shown to improve the training of PINNs \cite{wang2022and} and PI-DeepONets \cite{wang2022improved}. While a detailed discussion of the NTK is outside the scope of this work, we include it as implemented in \cite{wang2022improved} as an example of combining stacking physics-informed training with existing methods for improving physics-informed training. 

To use NTK weights, we begin by rewriting the loss function in Eq. \ref{eq:DeepONet_loss} as \cite{wang2022improved}:
\begin{equation}
        \mathcal{L}_{DoN}(\theta) = \frac{1}{N^*}\sum_{i=1}^{N^*}\left[ \mathcal{T}^{(i)}\left(\mathbf{u}^i, \mathcal{G}_\theta(\mathbf{u}^i)(\mathbf{x}^i)\right) \right]^2
\end{equation}
where $R=R_{bc}$, $N^* = 2NR$, and $\mathcal{T}^{(i)}$ denotes the operators in the loss function, including the boundary condition and differential operator. The NTK matrix is found as    
\begin{equation}
    H_{ij}(\theta) = \left\langle 
    \frac{d\mathcal{T}^{(i)}(\mathbf{u}^i, \mathcal{G}_\theta(\mathbf{u}^i)(\mathbf{x}_i)}{d\theta},     \frac{d\mathcal{T}^{(j)}(\mathbf{u}^j, \mathcal{G}_\theta(\mathbf{u}^j)(\mathbf{x}_j)}{d\theta}
    \right\rangle.
\end{equation}
Then, we can define the NTK weights at a given iteration $n$ by
\begin{equation}
    \lambda_k = \left( \frac{\max_{1\leq k\leq N^*} H_{kk}(\theta_n)}{H_{kk}(\theta_n)} \right)^\alpha
\end{equation}
where we take $\alpha = 0.5.$ The loss function with NTK weights incorporated is
\begin{equation}
        \mathcal{L}_{DoN, NTK}(\theta) = \frac{1}{N^*}\sum_{i=1}^{N^*}\lambda_k\left[ \mathcal{T}^{(i)}\left(\mathbf{u}^i, \mathcal{G}_\theta(\mathbf{u}^i)(\mathbf{x}^i)\right) \right]^2.
\end{equation}
We will give an example of the use of the NTK with the stacking PI-DeepONets in Sec. \ref{sec:deeponet}.

\section{Results with PINNs}\label{sec:PINN} 
We first demonstrate the performance of stacking networks by exploring some cases where PINNs are known to fail to train. In particular, we consider the damped pendulum problem in Sec.~\ref{sec:pendulum}, followed by a toy multiscale problem in Sec.~\ref{sec:1st_deriv_multiscale}. After that, we thoroughly investigate solving the 1D wave equation by the means of PINNs in Sec.~\ref{sec:wave}. In particular, we consider the cases where the same equation is enforced at all stacking levels as well as where different equations are imposed at different levels to further illustrate the flexibility of the stacking idea. Finally, numerical experiments using PI-DeepONets are shown in Sec.~\ref{sec:deeponet}.

\subsection{Pendulum} \label{sec:pendulum}
\begin{figure}[h!]
    \centering
    \begin{subfigure}{\textwidth}
    \centering
            \includegraphics[width=0.7\textwidth]{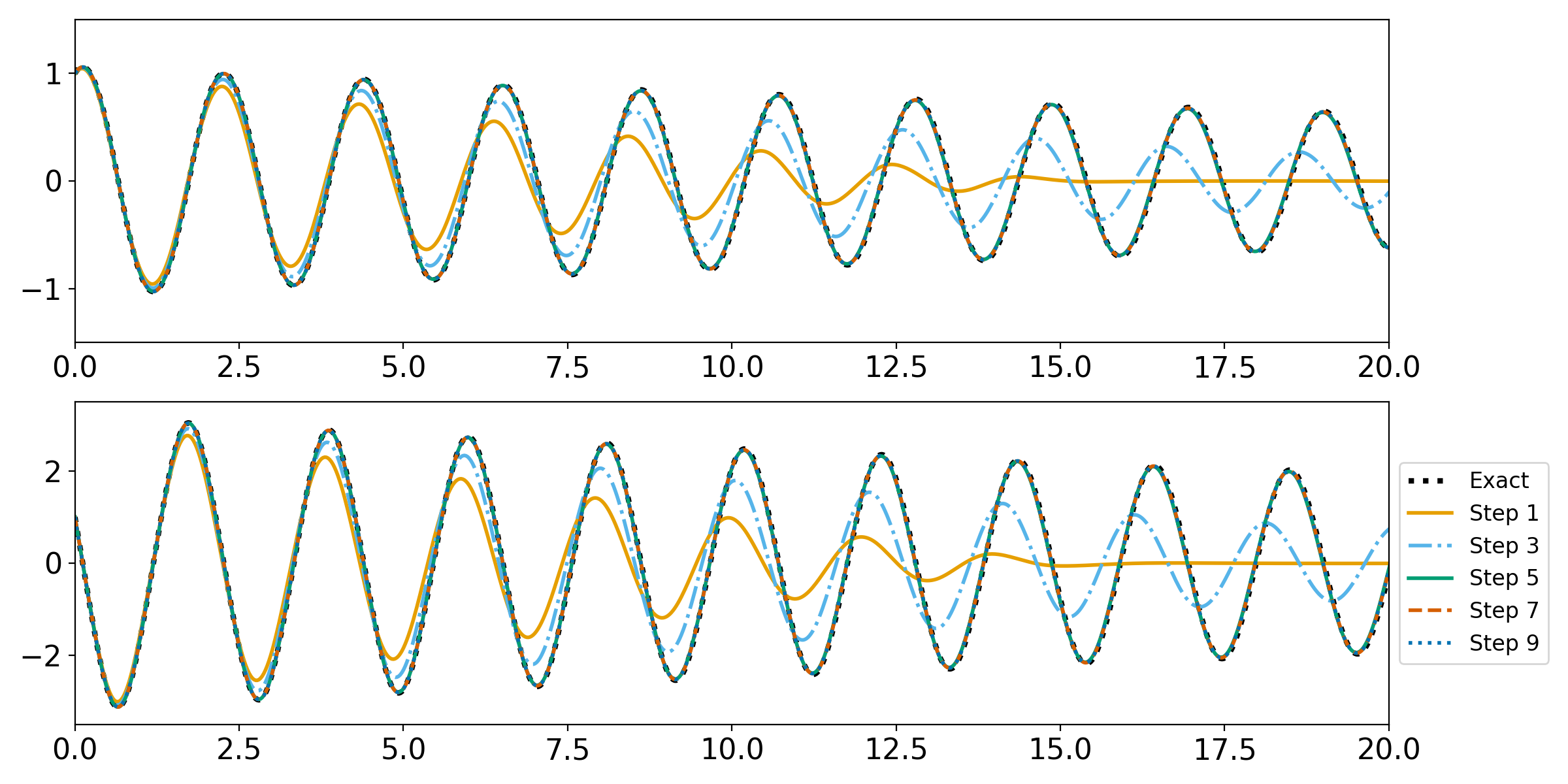}
                    \caption{Stacking PINN results for an illustrative example of $s_1$ (left) and $s_2$ (right) as a function of time for the pendulum problem up to nine stacking steps.}
    \end{subfigure}\hspace{0.02\textwidth}
    \begin{subfigure}{\textwidth}    \centering
            \includegraphics[width=0.45\textwidth]{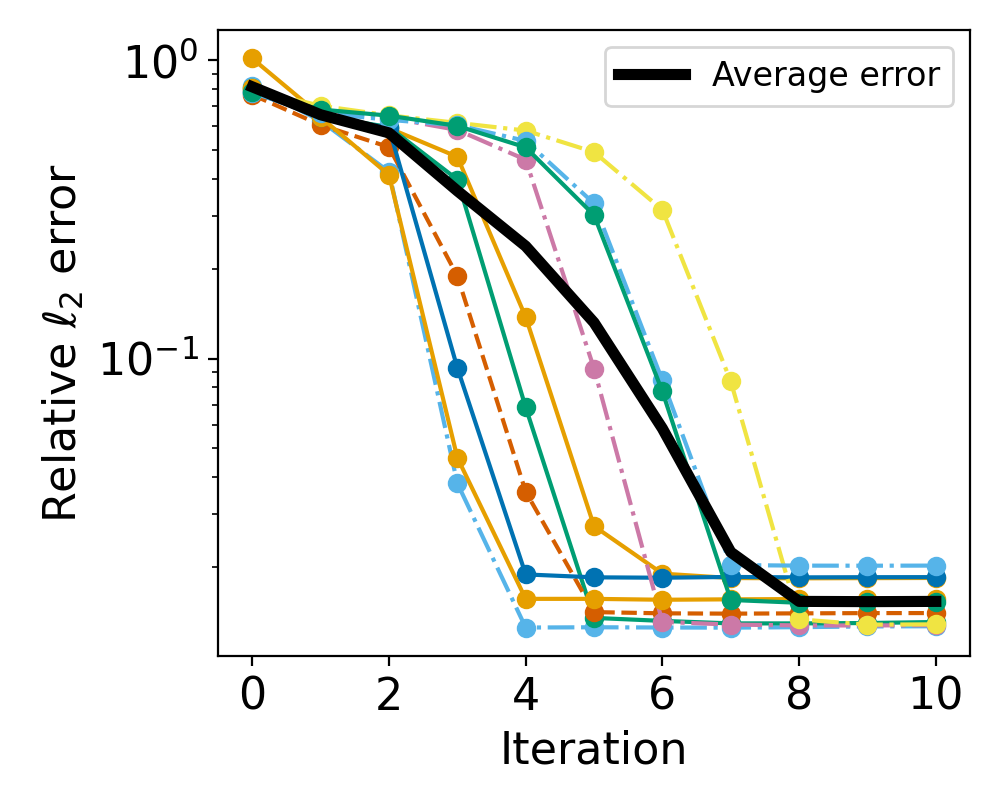}
        \caption{Pendulum relative $\ell_2$ training errors after each stacking step for ten initial seedings.}
    \end{subfigure}
    \caption{Stacking PINN training for the pendulum problem.}
    \label{fig:pendulum_stacking}
\end{figure}

To show how the method proposed in Sec. \ref{sec:method_stacking} works, we return to the pendulum case in Sec. \ref{sec:pinn_fails}. To set up the problem, we first train a single fidelity PINN for the pendulum problem, as in Fig. \ref{fig:SF_pen}. Clearly, this single fidelity PINN does not accurately capture the dynamics of the pendulum for long times. For each of the ten random initializations in Fig. \ref{fig:SF_pen}, we iteratively train a stacking PINN for up to ten stacking steps. After each stacking step, the solution gets closer to the exact solution for the pendulum, shown in Fig. \ref{fig:pendulum_stacking}a.

We define the relative $\ell_2$ error by
\begin{equation}
    error = \frac{||\mathcal{F}^i(x, t; \theta^i) - s(x, t)||_2}{||s(x, t)||_2}.
\end{equation}
After four to eight stacking steps, the pendulum reaches a fixed value of the relative $\ell_2$ error, which does not decrease further with additional stacking PINNs. The relative $\ell_2$ errors after each stacking iteration are shown in Fig. \ref{fig:pendulum_stacking}b.
The final relative errors are close to constant, and the initial relative errors are also quite close. Interestingly, each case seems to hit a critical point after which the relative $\ell_2$ error begins to decrease rapidly, before reaching and plateauing at the final value. 

For the damped pendulum, stacking PINNs allows for accurate solutions up to $T=20$.

\subsection{Multiscale problems} \label{sec:1st_deriv_multiscale}
We now consider a toy model, inspired by \cite{moseley2023finite}:
\begin{align}
    \frac{ds}{dx} &= \omega_1\cos (\omega_1 x) + \omega_2\cos(\omega_2 x),\\
    s(0) &= 0,
\end{align}
on domain $\Omega = [0, 20]$ with $\omega_1 = 1$ and $\omega_2 = 15$. The exact solution for this problem is $s(x) = \sin(\omega_1 x) + \sin(\omega_2 x)$.

We consider several cases. This problem is possible to solve with a standard single fidelity (SF) PINN, however, it requires a very large network. In Table \ref{tab:error_multiscale}, we show that the stacking PINN can reach a relative $\ell_2$ error lower than the best single fidelity PINN with just three stacking levels, and with less than one third the the number of trainable parameters needed in the single fidelity case. Importantly, the stacking PINN continues to improve with more multifidelity steps, and after ten stacking levels it reaches a relative error an order of magnitude lower than the best SF PINN, even though it has 61\% of the trainable parameters. 

The advantage of stacking PINNs for this problem is thus twofold. First, the stacking networks can reach a similar relative $\ell_2$ error to the relative $\ell_2$ error from a single fidelity network, but with a significantly smaller number of trainable parameters. For large applications pushing the memory limits of contemporary GPUs, this can offer an advantage where the network can be trained with a series of consecutive smaller networks, each of which is easier to train and does not have memory limitations. The cost of such training is, of course, the necessary wall clock time for sequential training of the networks. The second advantage of the stacking PINNs is that by adding additional levels, they can reach a significantly smaller relative $\ell_2$ error than the single fidelity PINN, while still possibly having fewer trainable parameters.

\begin{table}[h]
    \centering
    \begin{tabular}{||c | c | c | c ||} 
     \hline
     Method & Network size & Trainable parameters & Final relative error \\ 
     \hline
     Single fidelity & 3$\times$32 & 2209 & 1.3419 \\ 
      Single fidelity & 4$\times$64 & 12673 & 0.6543 \\ 
      Single fidelity & 4$\times$128 & 49921 & 0.1480 \\ 
      Single fidelity & 5$\times$64 & 16833 &  0.0949 \\ 
    %  SF, SA & 3$\times$32 & 2209 & 0.6172 \\ 
    % SF, SA & 4$\times$64 & 12673 & 1.2707 \\ 
    % SF, SA & 4$\times$128 & 49921 & 0.2100 \\ 
    %SF, SA & 5$\times$64 & 16833 &  0.2014 \\ 
     Stacking & 4$ \times$ 16, 1$\times$ 5, 3 stacking levels & 4900 & 0.0249 \\ 
     Stacking & 4$ \times$ 16, 1$\times$ 5, 10 stacking levels & 11179 & 0.0061 \\ 
     %Stacking, SA & 4$ \times$ 16, 1$\times$ 5, 5 stacking levels & 6694 & 0.0705\\
     %Stacking, SA & 4$ \times$ 16, 1$\times$ 5, 10 stacking levels & 11179 & 0.0202 \\
     \hline
    \end{tabular}
    \caption{Relative $\ell_2$ errors and network sizes for the multiscale problem in Sec. \ref{sec:1st_deriv_multiscale}. All stacking PINNs begin with a single fidelity network with size 3$\times$32. For the stacking cases, we report the first stacking level that has a relative $\ell_2$ error less than the minimum single fidelity error, and the relative $\ell_2$ error after the tenth stacking level.}
    \label{tab:error_multiscale}
\end{table}

\begin{figure}[h]
    \centering
    \begin{subfigure}{0.48\textwidth}
            \includegraphics[width =\textwidth]{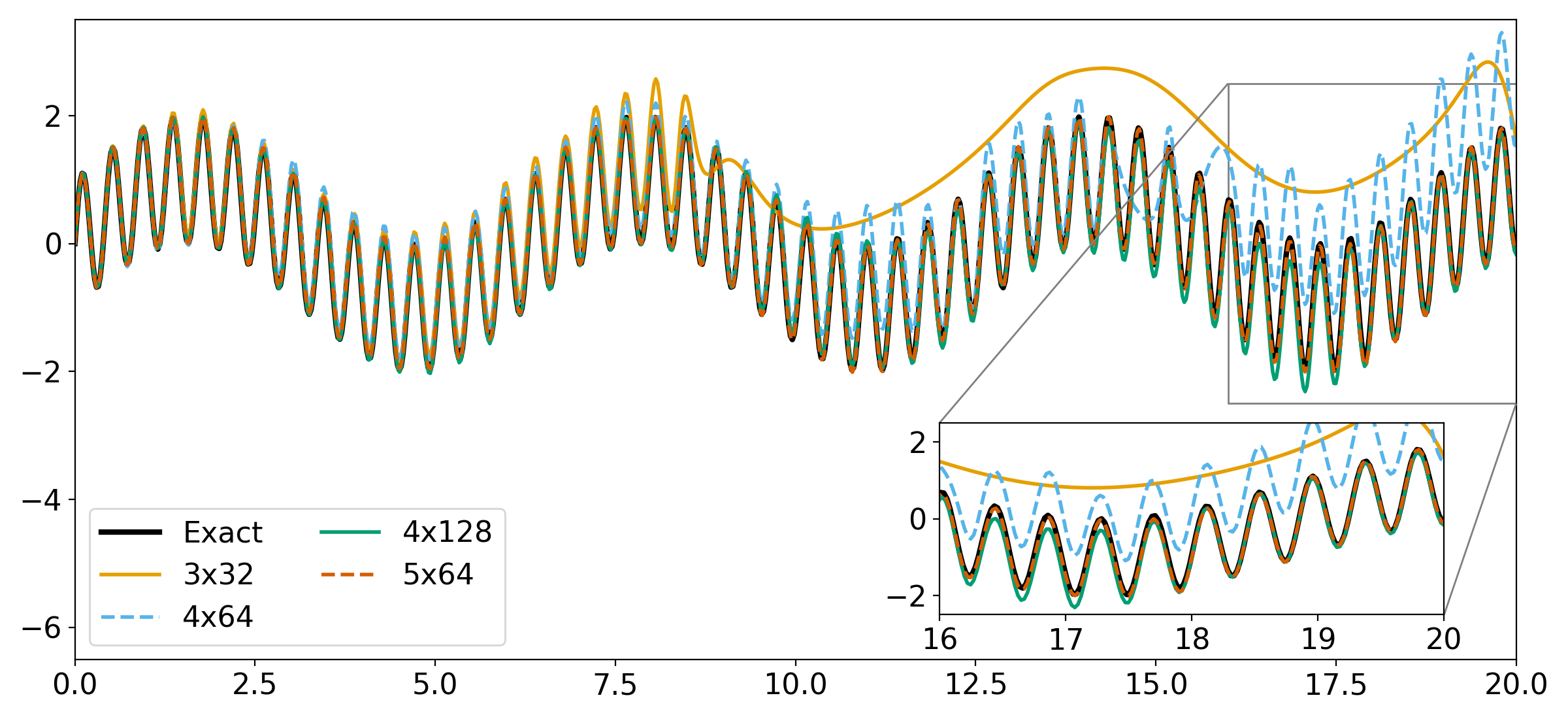}
        \caption{Single Fidelity}
    \end{subfigure}
    \begin{subfigure}{0.48\textwidth}
            \includegraphics[width =\textwidth]{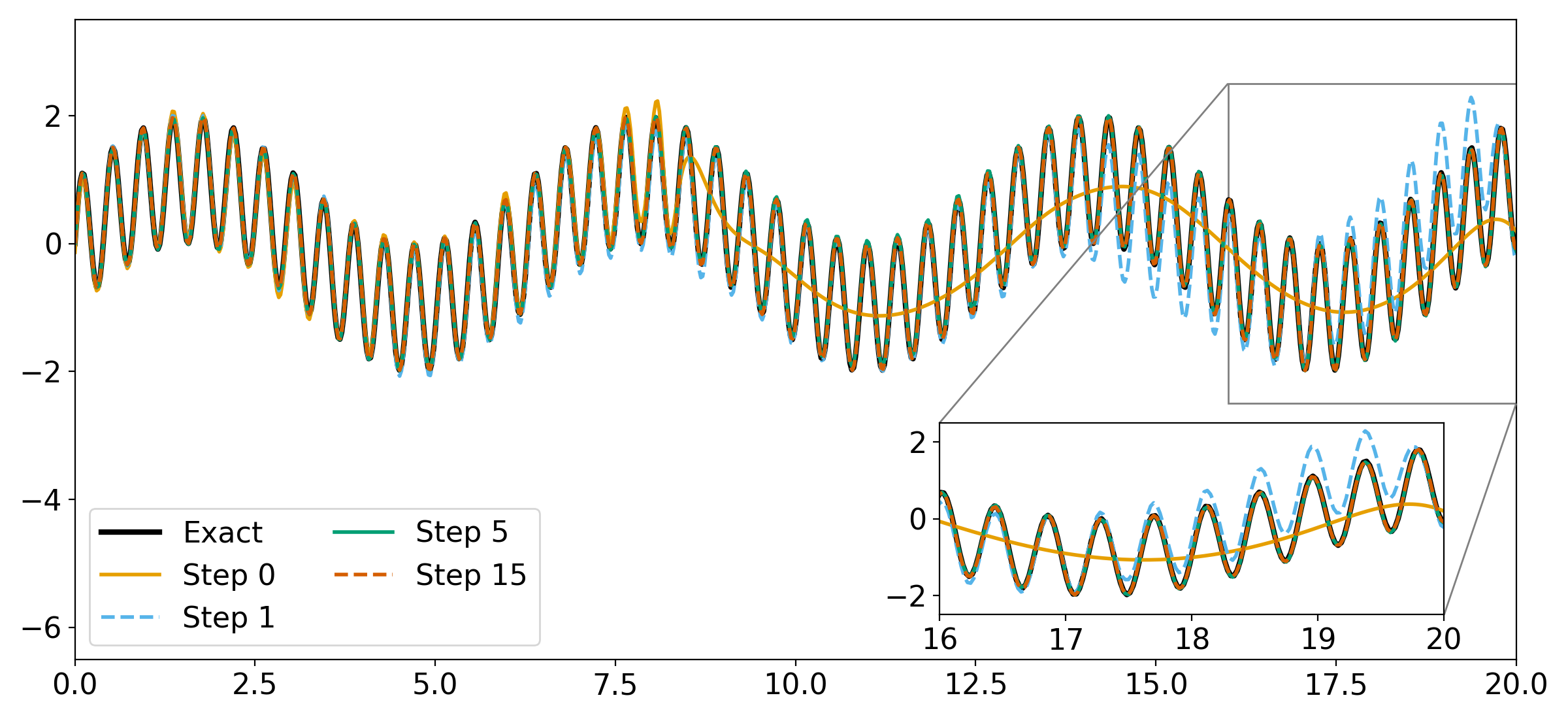}
        \caption{Stacking PINN}
    \end{subfigure}
   \begin{subfigure}{0.48\textwidth}
   \centering
            \includegraphics[width = 0.5\textwidth]{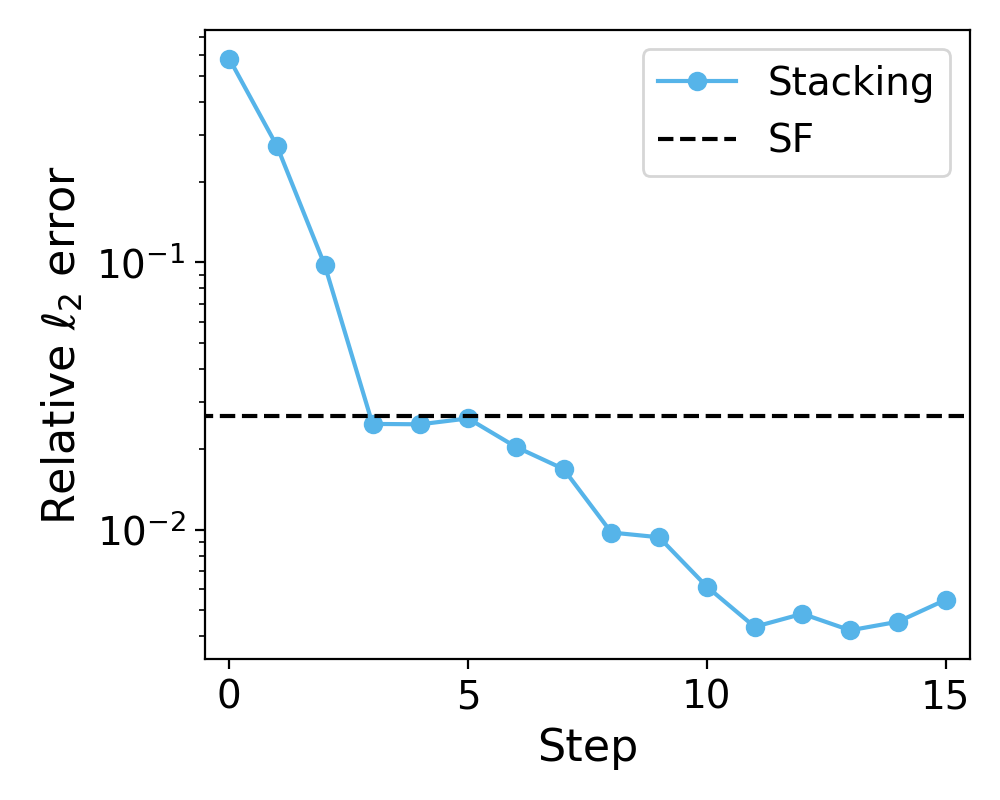}
       \caption{Relative $\ell_2$ errors}
   \end{subfigure}
    \caption{Results for the multiscale problem. (a) Single fidelity results for a variety of network sizes. (b) Multifidelity stacking results. %(c) Multifidelity stacking results {\it with} self-adaptive weights. 
    (c) Relative $\ell_2$ errors for the stacking PINN. The black dashed line in (c) is the lowest error from the single fidelity training for the $5\times 64$ network. For the stacking PINN, step 0 is the single fidelity step and step 1 is the first multifidelity step.}
    \label{fig:Multiscale}
\end{figure}

\subsection{Wave equation}\label{sec:wave}
The wave equation is given by 
\begin{align}
    s_{tt}(x, t) -c^2 s_{xx}(x, t) = 0, \quad &x\in(0, 1)\times(0, 1), \label{eq:wave1} 
\end{align}
where $c$ is a scalar parameter for wave speed. We consider the following boundary and initial conditions:
\begin{align}
    s(0, t) = s(1, t)  = 0, \quad &t\in [0, 1], \label{eq:wave2} \\
    s(x, 0) = \sin(\pi x) + a \sin(4 \pi x), \quad &  x \in  [0, 1], \label{eq:wave3} \\
    s_t(x, 0) = 0, \quad &x \in [0, 1].\label{eq:wave4} 
\end{align}

The exact solution for the wave equation is given by 
\begin{equation}
    s(x, t) = \sin(\pi x)\cos(2 \pi t) + a \sin(4 \pi x)\cos(4 c \pi t). \label{eq:wave_exact}
\end{equation}

We use the wave equation to illustrate another key advantage of the stacking PINN framework. While up to this point we have trained with the same equations for each stacking level, it is possible, and in some cases advantageous, to change the equations for each stacking level to make it easier to train the early levels. We have found that a well-trained low-fidelity model, possibly for a slightly different and simpler equation, can produce more robust results than training with the same equation for all stacking levels. One advantage of training with an equation that is easier to learn with the single fidelity network is that one can generally use a smaller network, limiting the number of training parameters.

\begin{table}
    \centering
    \begin{tabular}{||c | l | c||} \hline
        Case & Schedule for $c$ & Error \\ \hline 
        Case 1 & $c=2$ for all stacking iterations and the single fidelity network & $2.913 \times 10^{-3}$ \\
        Case 2 & $c = [1, 1.25, 1.5, 1.75, 2]$, and then $c=2$ for all remaining iterations & $4.274 \times 10^{-3}$ \\
        Case 3 & $c=1$ for the single fidelity level and $c=2$ for all stacking iterations  & $3.805 \times 10^{-3}$\\
        Case 4 & $c = [1.0, 2.0, 2.0, 3.0, 3.0,3.0,3.0]$, and then $c=4$ for remaining iterations  & $8.396 \times 10^{-3}$\\ 
        Single fidelity & $c = 2$ & $4.129\times 10^{-2}$ \\
        Single fidelity & $c = 4$ & 1.379\\ \hline
    \end{tabular}
    \caption{Schedule for increasing $c$ with stacking iterations for each of the cases considered for the wave equation case. The right column has the relative $\ell_2$ error for the final stacking layer. Note that this is calculated with respect to the exact solution with $c=2$ for cases 1, 2, and 3, and $c=4$ for Case 4.}
    \label{tab:c_variations}
\end{table}
For the wave equation, we consider four cases, varying the parameter $c$ in Eq. \ref{eq:wave1} during each stacking level according to Table \ref{tab:c_variations}. When $c$ is smaller, the PINN is much easier to train. This, by starting $c$ small and increasing to our target value of $c=2$ or $c=4$, we hope to begin with a very accurate prediction, for a slightly modified equation, and then correct with the prediction for the correct equation. 

\begin{figure}[h]
    \centering
            \includegraphics[width=0.8\textwidth]{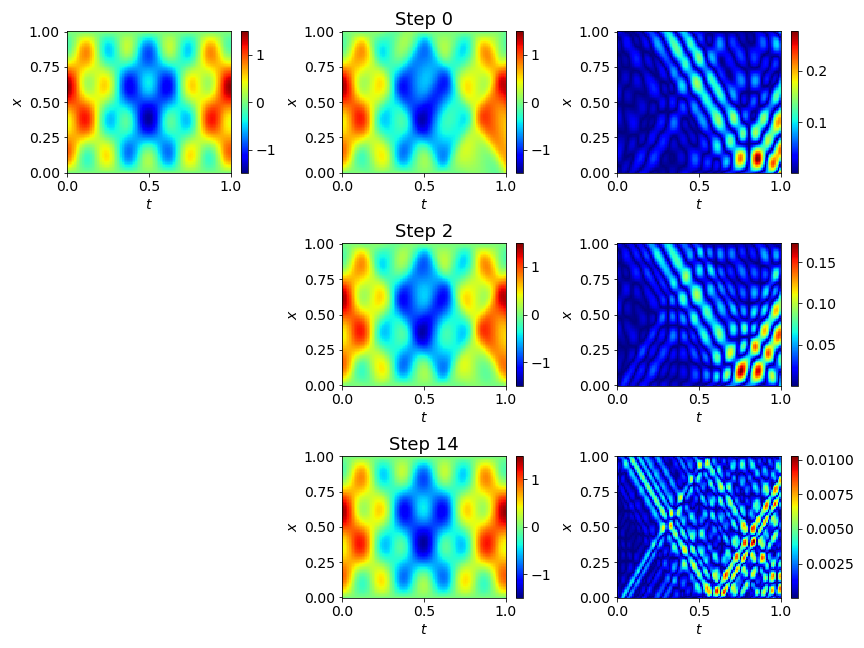}
    \caption{Results for the stacking PINN for the wave equation for Case 1. The top left figure is the exact solution. The results from training steps 0, 2, and 14 are in the middle column. Their respective absolute errors are in the right column.}
    \label{fig:wave}
\end{figure}

\begin{figure}[h]
    \centering
            \includegraphics[width=0.8\textwidth]{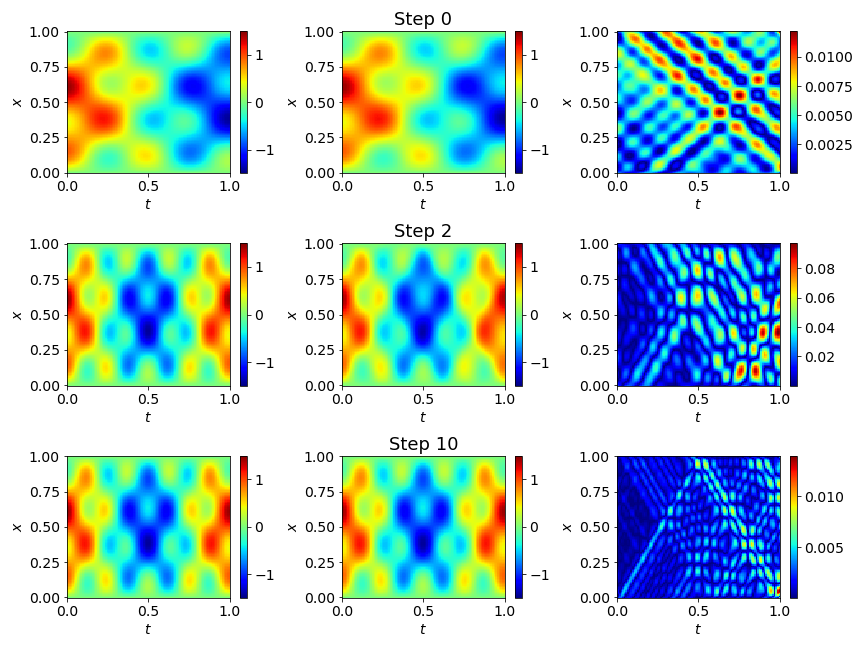}
    \caption{Results for the stacking PINN for the wave equation for Case 3. The left column has the exact solution for the value of $c$ used for a given stacking step. The results from training steps 0, 2, and 10 are in the middle column. Their respective absolute errors are in the right column. }
    \label{fig:wave_c2}
\end{figure}

\begin{figure}[h]
    \centering
            \includegraphics[width=0.4\textwidth]{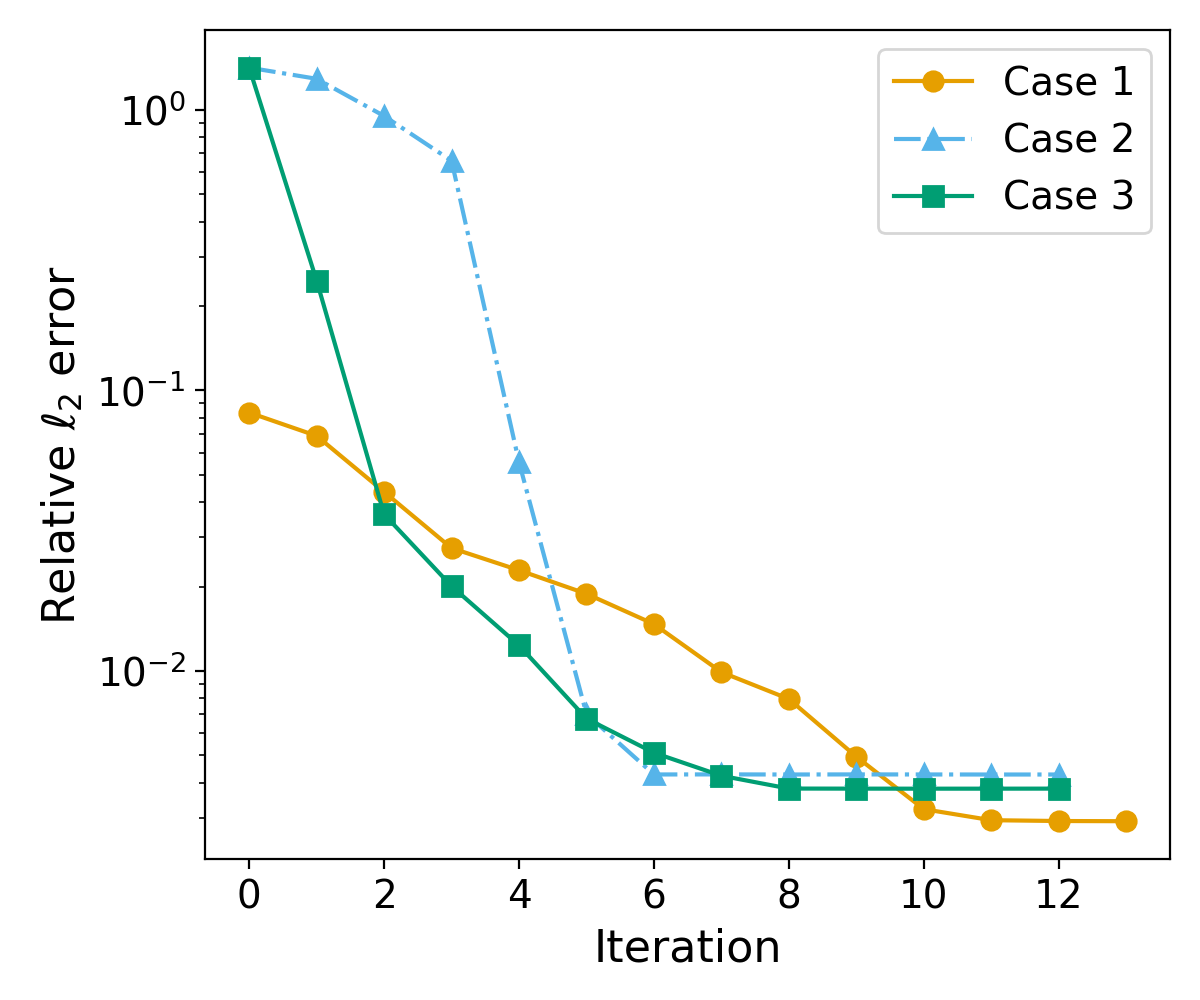}
    \caption{Relative $\ell_2$ errors for the stacking PINNs for the wave equations for Case 1 (circles), Case 2 (triangles), and Case 3 (squares). The relative $\ell_2$ is calculated using the exact solution for the wave equation with $c=2$.}
    \label{fig:wave_errors}
\end{figure}

In Fig. \ref{fig:wave}, we show the training for Case 1, where $c$ is held fixed at $c=2.$ The final relative $\ell_2$ error is $2.913 \times 10^{-3}$ after 14 stacking levels. When compared with Case 3 where $c$ is varied, shown in Fig. \ref{fig:wave_c2}, Case 1 reaches a lower relative $\ell_2$ error after many stacking layers, see Fig. \ref{fig:wave_errors}. However, Case 3 converges to a smaller relative $\ell_2$ error in early iterations. Depending on the performance goal, Case 3 may represent a need for less training through using fewer stacking levels, while still achieving accurate results. The final relative $\ell_2$ error for Case 3 is $3.805 \times 10^{-3}$. Case 2 is shown in Appendix \ref{sec:appendix_wave}, and reaches a final relative $\ell_2$ error of $4.274 \times 10^{-3}$. 

Cases 1, 2, and 3 have relatively similar training performance as shown by similar values of the final relative $\ell_2$ errors, and all result in accurate predictions for the wave equation with a one order of magnitude reduction in the error compared to the single fidelity PINN. Case 4 represents a case of pushing to accurately predict a more difficult problem, with $c=4$. The stacking PINN is able to accurately capture the fine-scale oscillations of this problem, shown in Figs. \ref{fig:wave_c3_app} and \ref{fig:wave_c3errors_app} in Appendix \ref{sec:appendix_wave}, with a relative $\ell_2$ error of $8.396\times 10^{-3}$ for the final stacking layer. This relative $\ell_2$ error is three orders of magnitude smaller than the single fidelity training for the wave equation with $c=4.$

\section{Results with DeepONets}\label{sec:deeponet} 
While a great deal of work has been put into physics-informed DeepONets \cite{wang2021learning, wang2022improved, wang2023long}, there is still room for improving training for complex physics-informed operator networks. We present results that show that the stacking method can be applied to physics-informed training for DeepONets. 

\begin{figure}[h]
    \centering
            \includegraphics[width=0.8\textwidth]{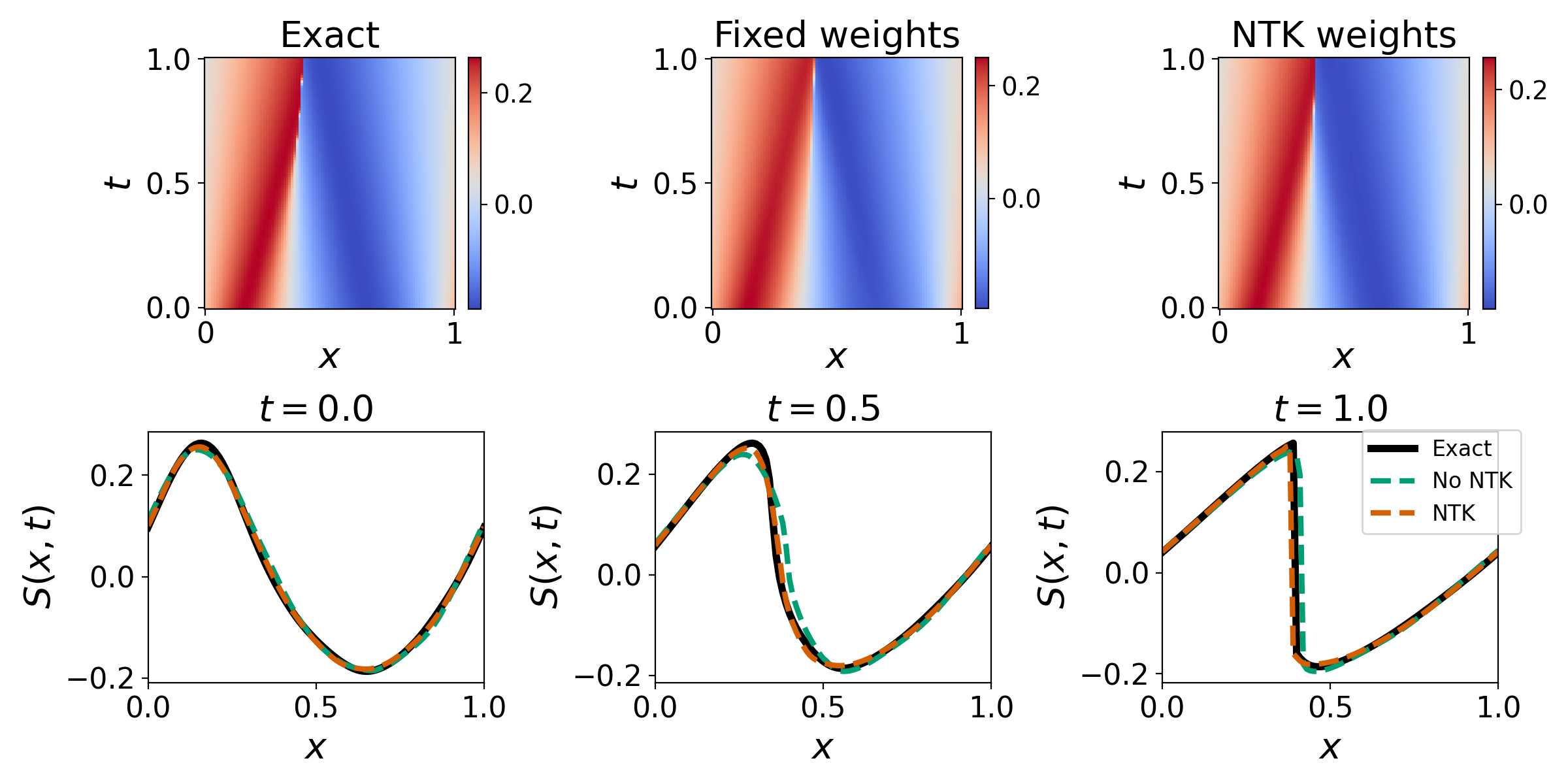}
    \caption{Single fidelity DeepONet results for a random initial condition not seen during training. The exact solution is in the top left. Training with fixed weights is in the top middle column, and training with NTK weights is in the top right column. The bottom has slices of the solution at $t=0, 0.5$, and $1$. Clearly, the addition of the NTK weights improves the accuracy of the results.  }
    \label{fig:SF_DeepONet}
\end{figure}

We consider the test case for the viscous one-dimensional Burgers equation with periodic boundary conditions presented in \cite{wang2022improved, wang2021learning} and \cite{howard2023multifidelity}. The equations are given by
\begin{align}
  \textcolor{black}{  \frac{\partial s}{\partial t} + s \frac{\partial s}{\partial x} - \nu \frac{\partial ^2s}{\partial x^2}} &=    0, \; (x, t) \in [0, 1] \times [0, 1] \\
  s(x, 0) &= u(x), \; x\in[0, 1], \\
  s(0, t) &= s(1, t), \; t\in[0, 1], \\
  \frac{\partial s}{\partial x}(0, t) &= \frac{\partial s}{\partial x}(1, t), \; t\in[0, 1],
\end{align}
where $\nu$ is the viscosity. We generate initial conditions $u(x)$ from a Gaussian random field $\sim \mathcal{N}(0, 25^2(-\Delta + 5^2I)^{-4})$. The initial conditions are sampled at $P_{IC} = 101$ uniformly spaced locations on $x \in [0, 1]$, and the boundary conditions are randomly sampled at $P_{BC} = 100$ locations on $(x, t) = (0, t)$ and $(x, t) = (1, t)$. The residual is evaluated on $P_p = 2,500$ randomly sampled collocation points from the interior of the domain. We train with $N=1000$ samples of the initial condition $u(x)$. The performance is tested by generating an additional $N_{test} = 100$ initial conditions and simulating the solution with Matlab using the Chebfun \cite{Driscoll2014} package following the method in \cite{wang2022improved}.

This problem has been shown to be especially difficult to train for small viscosity $\nu = 0.0001$, which we take as our target viscosity \cite{wang2021learning, wang2022improved}. Single fidelity training results are shown in Fig. \ref{fig:SF_DeepONet}. The DeepONet struggles to capture the areas that form near the shock with steep gradients. The NTK does significantly improve the accuracy of the results, however, it is very computationally intensive to compute, increasing the training time by about a factor of three. 

We train four cases using stacking DeepONets, with and without the NTK and with fixed and changing viscosity values. When the viscosity is fixed, we use $\nu = 0.0001$ for each training step. When the viscosity is changed, we train the SF DeepONet with $\nu = 0.01$, then reduce $\nu$ by a factor of ten for each stacking step to reach $\nu = 0.0001$. These cases are shown in Table \ref{tab:DeepONet_cases}.

\begin{table}
    \centering
    \begin{tabular}{||c | c | c| c | c||} \hline
        Case & Schedule for $\nu$ & Weighting scheme & Stacking levels &  Error \\ \hline 
        Case 1 & Fixed & Fixed & 10 & $21.83\% \pm 10.10 \%$\\
        Case 2 & Changing & Fixed & 10 & $14.63 \% \pm 9.08\%$\\
        Case 3 & Fixed & NTK & 6 & $10.06 \% \pm 6.27 \%$ \\ 
        Case 4 & Changing & NTK & 6& $5.92\% \pm 4.88\%$\\ 
        Single fidelity & -- & Fixed & -- &$ 26.60\% \pm  13.52\%$ \\ 
        Single fidelity & -- & NTK & -- &$11.19\% \pm 6.36\%$  \\ \hline
    \end{tabular}
    \caption{Cases used in training for the stacking DeepONets for Burgers equation. A fixed schedule for $\nu$ denotes that $\nu = 0.0001$ for all stacking levels. A changing schedule denotes that $\nu = 0.01$ for Step 0, $\nu = 0.001$ for Step 1, and then $\nu = 0.0001$ for all remaining stacking levels. The reported error is the relative $\ell_2$ error.}
    \label{tab:DeepONet_cases}
\end{table}

\begin{figure}[h]
    \centering
            \includegraphics[width=0.4\textwidth]{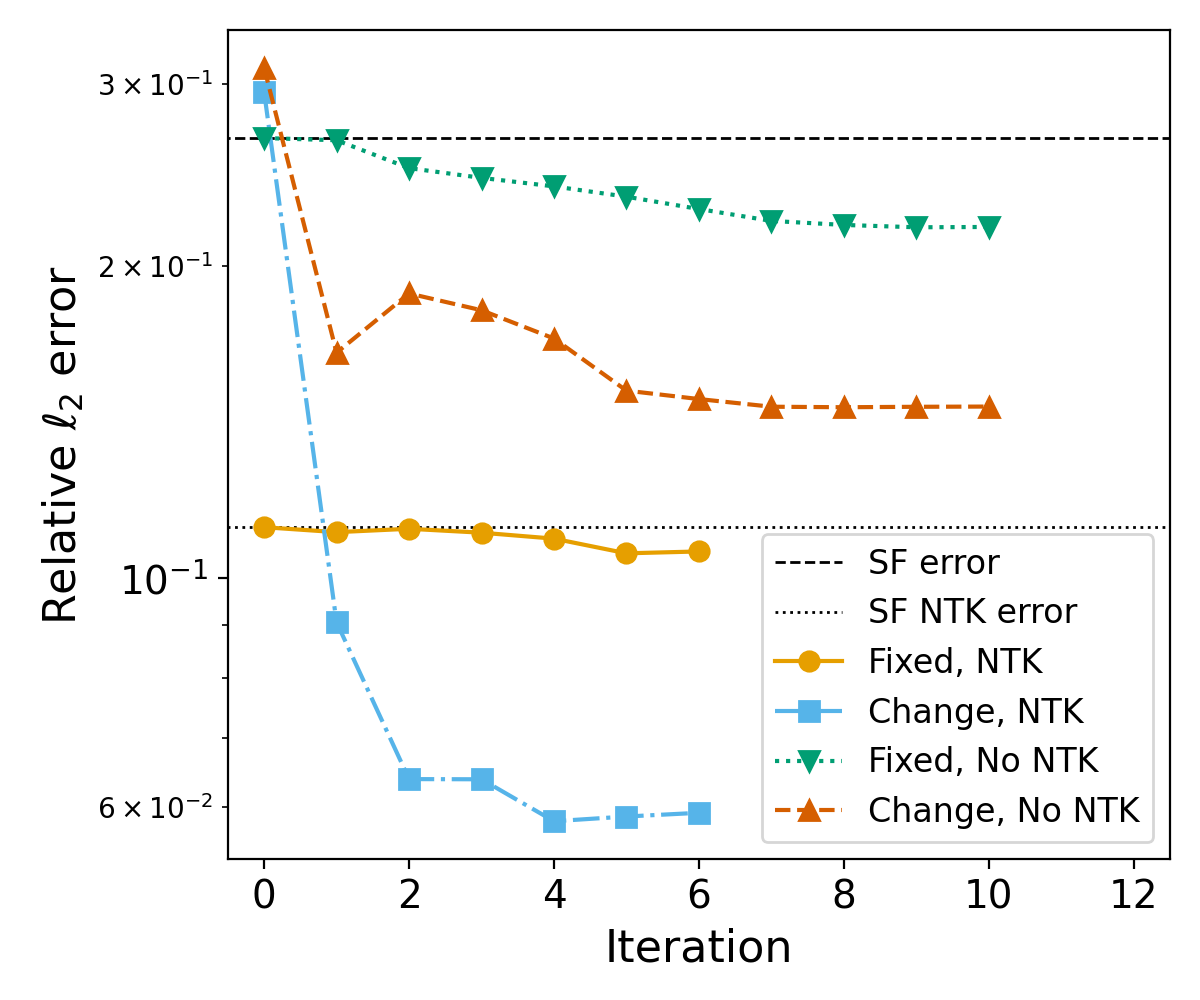}
    \caption{Evolution of the relative $\ell_2$ error for the four DeepONet cases considered. The relative $\ell_2$ error is calculated with respect to the numerically generated solution for 100 solutions in the test set for $\nu = 0.0001$. ``Fixed'' denotes cases where $\nu$ is fixed, and ``Change'' denotes cases where $nu$ starts large and is decreased. All errors are calculated with respect to the exact solutions with $\nu =0.0001$.}
    \label{fig:DeepONet_error}
\end{figure}

\begin{figure}[h]
    \centering
            \includegraphics[width=0.8\textwidth]{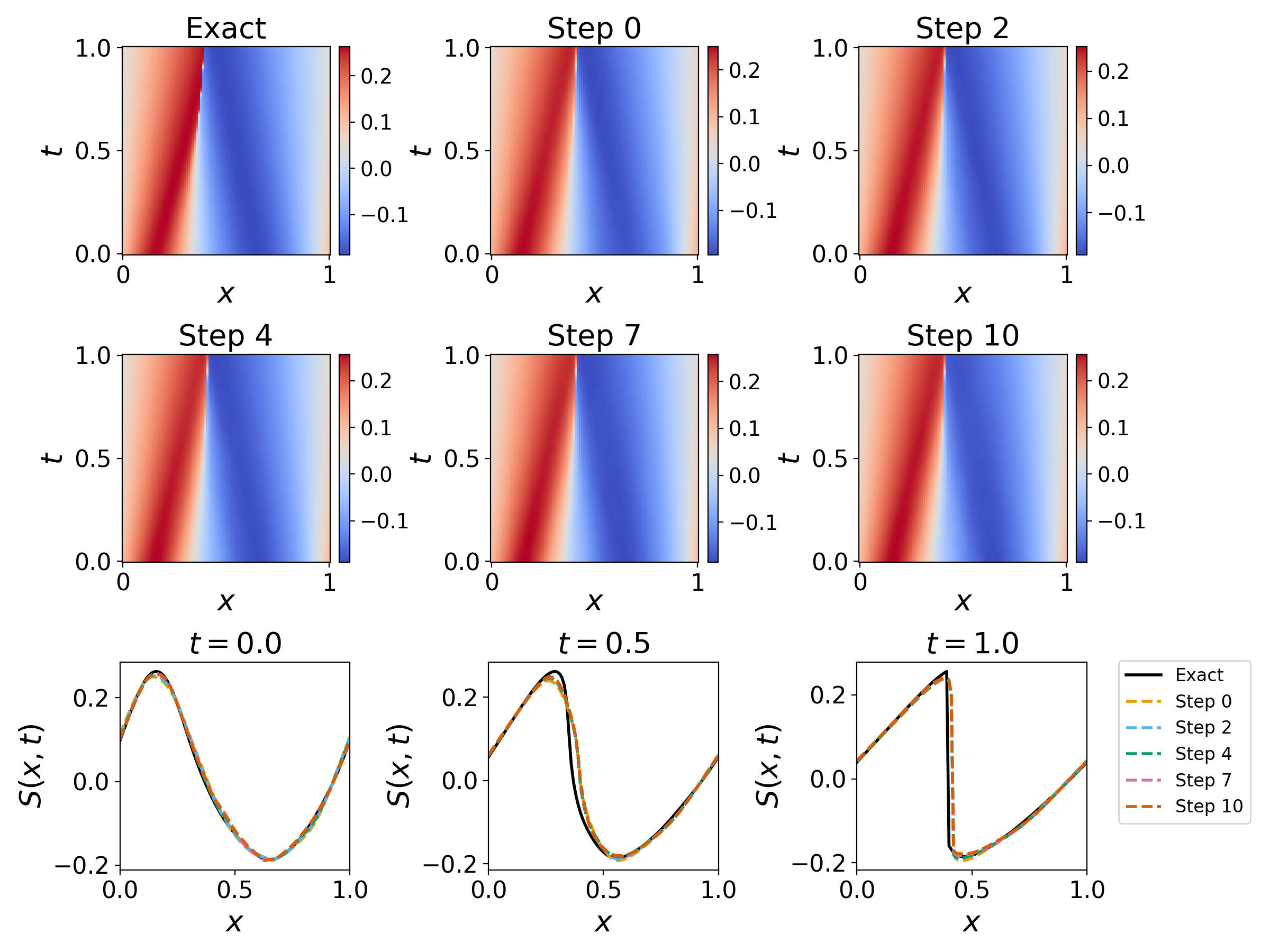}
    \caption{Results for a stacking DeepONet with fixed $\nu$ and fixed weights. Representative steps are shown from the training, including the final stacking step 10. While the results do improve from the initial predictions in early stacking steps, the results struggle in the vicinity of the sharp gradients.}
    \label{fig:DeepONet_fixednu_fixedweights}
\end{figure}

\begin{figure}[h]
    \centering
            \includegraphics[width=0.8\textwidth]{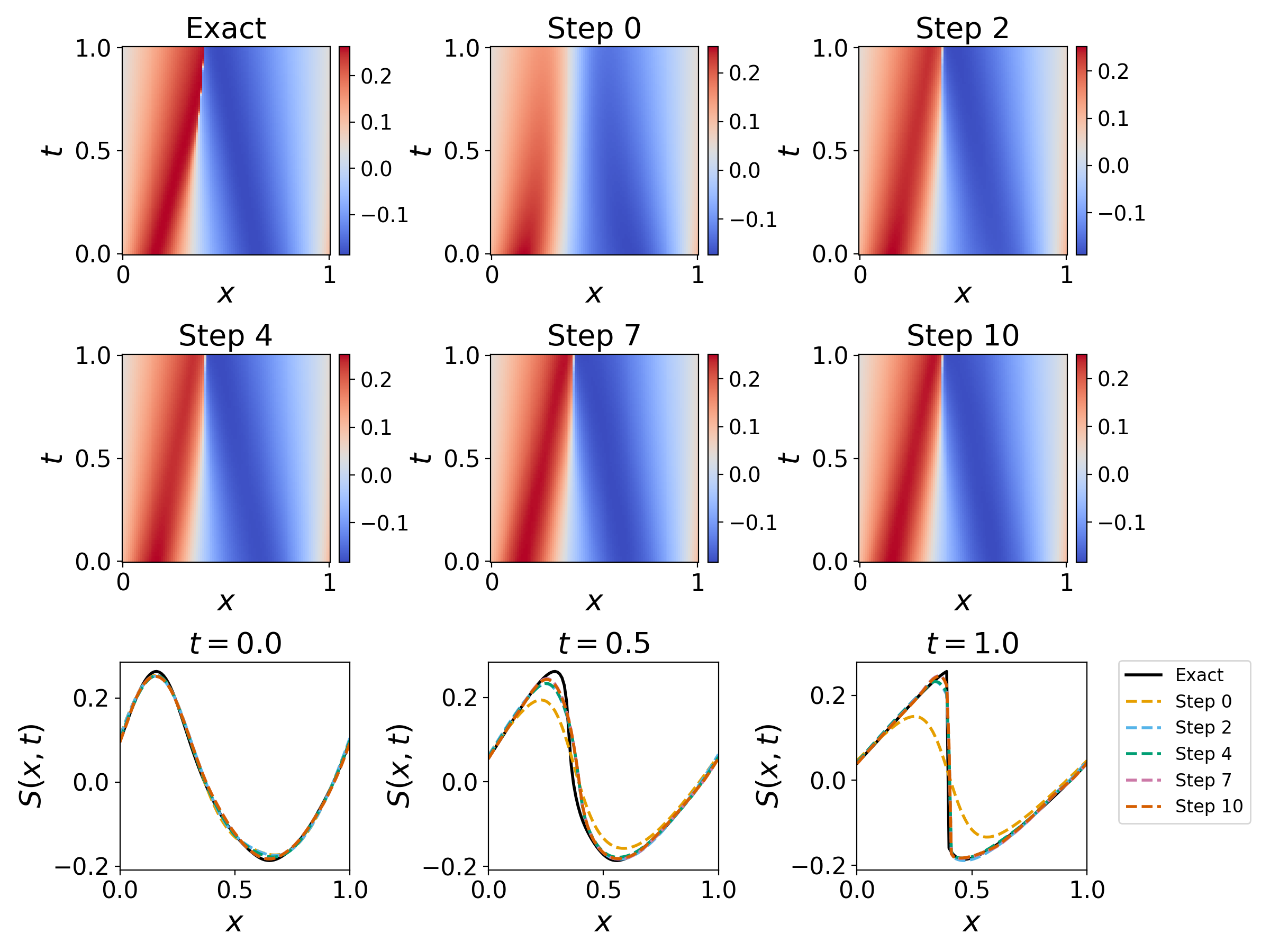}
    \caption{Results for a stacking DeepONet with changing $\nu$ and fixed weights. Representative steps are shown from the training, including the final stacking step 10.}
    \label{fig:DeepONet_changenu_fixedweights}
\end{figure}

From Table \ref{tab:DeepONet_cases}, the cases where $\nu$ changes by starting large and decreasing to the target value clearly have the best performance. When $\nu$ is fixed, the stacking DeepONets can struggle to train to a desired accuracy, however the performance is improved relative to the single fidelity DeepONet. While the relative $\ell_2$ error does slowly decrease with additional stacking levels with fixed $\nu$, shown in Fig. \ref{fig:DeepONet_error}, the time to train additional stacking levels is cost prohibitive when compared with the case where $\nu$ is gradually decreased, which show better performance both with and without the NTK. Comparing the case where $\nu$ is fixed, in Fig. \ref{fig:DeepONet_fixednu_fixedweights}, with the case where $\nu$ changes, in Fig. \ref{fig:DeepONet_changenu_fixedweights}, both with fixed weights, it is clear that changing $\nu$ is advantageous to reach a lower error with fewer stacking steps.

\begin{figure}[h]
    \centering
            \includegraphics[width=0.8\textwidth]{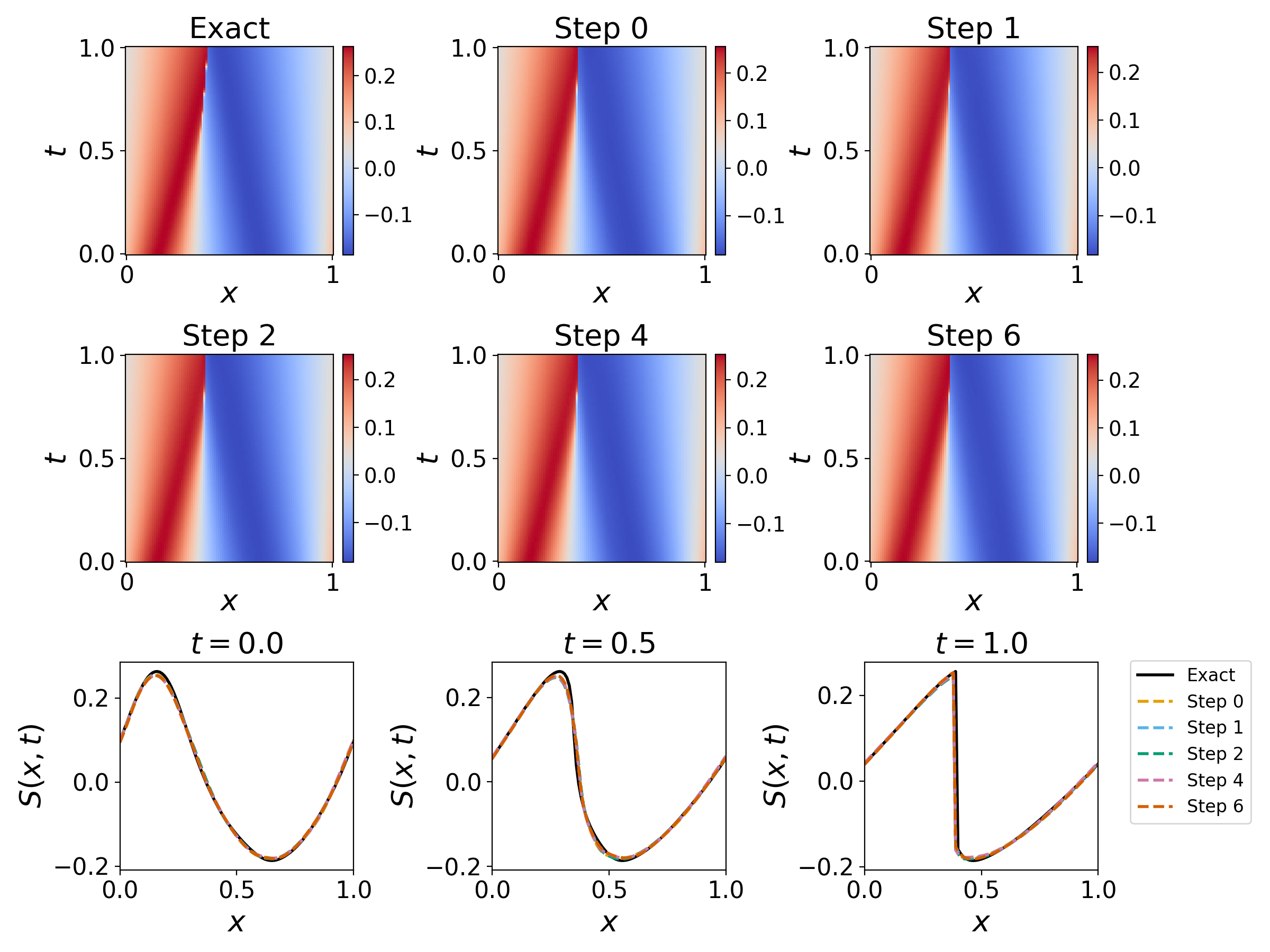}
    \caption{Results for a stacking DeepONet with fixed $\nu$ and NTK weights. Representative steps are shown from the training, including the final stacking step 6.}
    \label{fig:DeepONet_fixnu_NTKweights}
\end{figure}

\begin{figure}[h]
    \centering
            \includegraphics[width=0.8\textwidth]{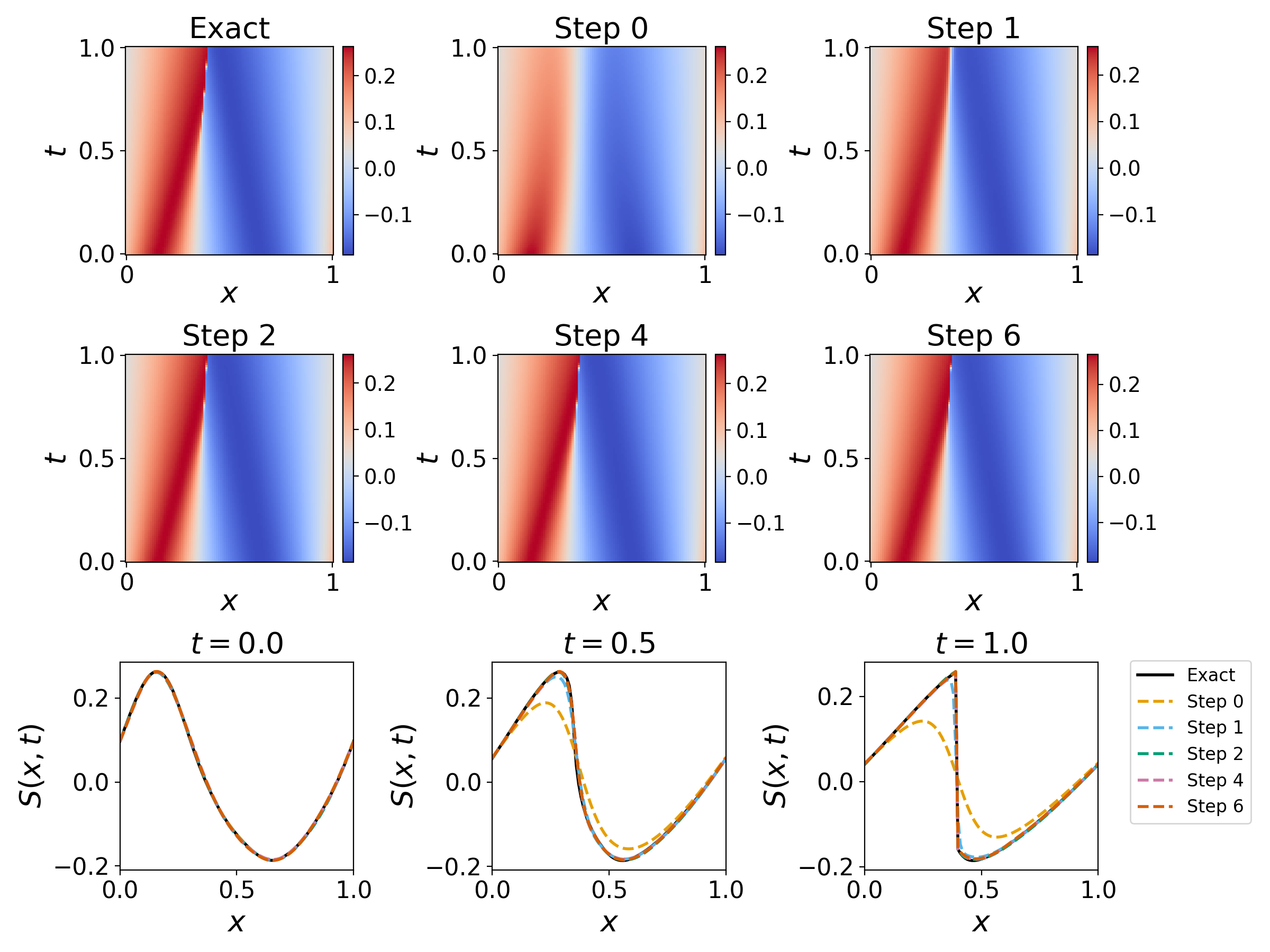}
    \caption{Results for a stacking DeepONet with changing $\nu$ and NTK weights. Representative steps are shown from the training, including the final stacking step 6.}
    \label{fig:DeepONet_changenu_NTKweights}
\end{figure}

Switching to NTK weights does improve the performance of the method, at the expense of greater computational cost. With the NTK, the case with fixed $\nu$, shown in Fig. \ref{fig:DeepONet_fixnu_NTKweights}, greatly improves. Most impressively, when $\nu$ starts large and decreases, shown in Fig. \ref{fig:DeepONet_changenu_NTKweights}, the final relative $\ell_2$ error decreases substantially compared to single fidelity training \cite{wang2022improved}. This suggests that stacking multifidelity DeepONets can be a key technique in addition to existing techniques to improve performance when physics-informed DeepONets fail to train.

\section{Discussion and conclusions} \label{sec:conclusion}
The inability of standard PINNs to train for some systems, even when given an extended amount of time to train, speaks to the need for the iterative stacking approach presented in this work. In this paper we have demonstrated how to train stacking networks to reach solutions for PINNs and PI-DeepOnets that can otherwise fail to train accurately. Through the use of stacking PINNs and PI-DeepONets, we are able to produce accurate solutions for long times and for equations where it is extremely difficult to train a standard PINN or PI-DeepONet.

Importantly, the method developed in this work can be used to boost further the performance of existing methods for improving training PINNs and PI-DeepONets. For example, it would lend itself nicely to methods for choosing adaptive residual points, and works well with the NTK (as shown in Sec. \ref{sec:deeponet}). 

In future work, there are remaining directions to explore. For example, while the multifidelity networks do not have to have the same sizes at each stacking step, in this work we chose the same size to enable transferring trained weights to each subsequent level instead of seeding randomly. To reduce the training cost, it may be advantageous to choose smaller networks for earlier stacking levels, which would provide a low fidelity prediction as in \cite{penwarden2022multifidelity}. Additionally, we have shown that for some problems it can be better to start the stacking networks with cases that are easier to learn, and then gradually change to the target problem. The rate at which the equation should change is an open problem, and likely target application specific. 

Importantly, the multifidelity method we presented here is not unique to feed-forward neural networks and DeepONets. Indeed, each block in the stacking setup could be replaced by different architectures. For instance, one could create a stacking convolutional neural network, or use a different operator learning method. This flexibility leads to a great deal of future work in studying optimal frameworks for physics-informed problems. Additionally, while keeping the number of layers in each stacking network fixed at each level allows for transferring trained weights to initialize new levels, for some applications increasing the network complexity by changing the architecture through adding width or depth to the network could lead to better training. We leave these as directions for future work. 

%\section{Code and data availability}
%Code and data needed for training the examples in this paper will be available  upon publication. 

\section{Acknowledgements}
      A. A. H. thanks Dr. Wenqian Chen for helpful discussions. S. H. M. acknowledges support from the National Science Foundation Mathematical Sciences Graduate Internship program. The work of S. E. A. is supported by the Department of Energy (DOE) Office of Advanced Scientific Computing Research (ASCR) through the Pacific Northwest National Laboratory Distinguished Computational Mathematics Fellowship (Project No. 71268). This project was completed with support from the U.S. Department of Energy, Advanced Scientific Computing Research program, under the Scalable, Efficient and Accelerated Causal Reasoning Operators, Graphs and Spikes for Earth and Embedded Systems (SEA-CROGS) project (Project No. 80278). The computational work was performed using PNNL Institutional Computing at Pacific Northwest National Laboratory. Pacific Northwest National Laboratory (PNNL) is a multi-program national laboratory operated for the U.S. Department of Energy (DOE) by Battelle Memorial Institute under Contract No. DE-AC05-76RL01830.

\bibliographystyle{unsrt}
\bibliography{references}  %%% Uncomment this line and comment out the ``thebibliography'' section below to use the external .bib file (using bibtex) .

\appendix

\section{Wave equation} \label{sec:appendix_wave}

In this section we show additional results for Case 2 and Case 4 of the wave equation example from Sec. \ref{sec:wave}.
\begin{figure}[h]
    \centering
            \includegraphics[width=0.8\textwidth]{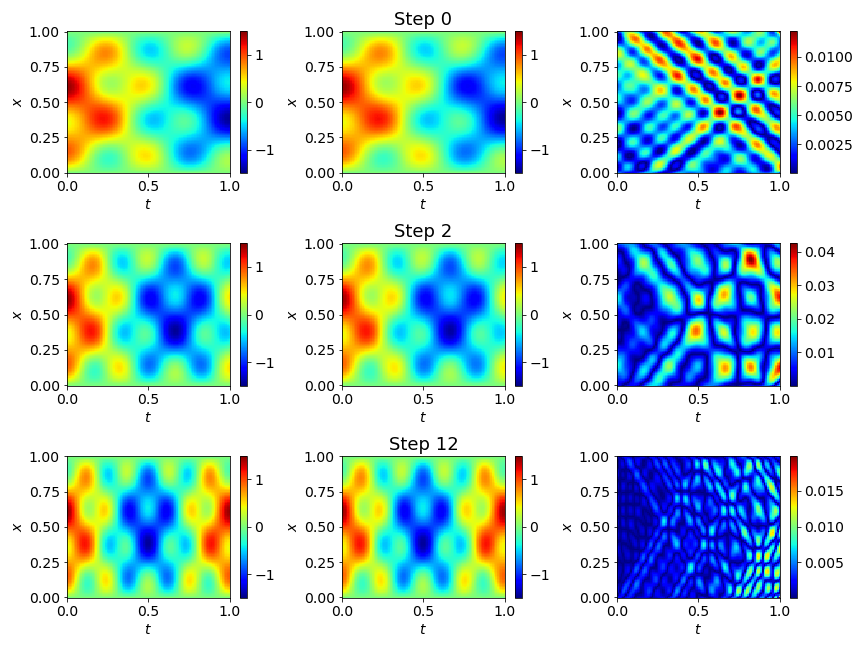}
    \caption{Results for the stacking PINN for the wave equation for Case 2. The left column has the exact solution for the value of $c$ used for a given stacking step. The results from training steps 0, 2, and 12 are in the middle column. Their respective absolute errors are in the right column. }
    \label{fig:wave_c2_app}
\end{figure}

\begin{figure}[h]
    \centering
            \includegraphics[width=\textwidth]{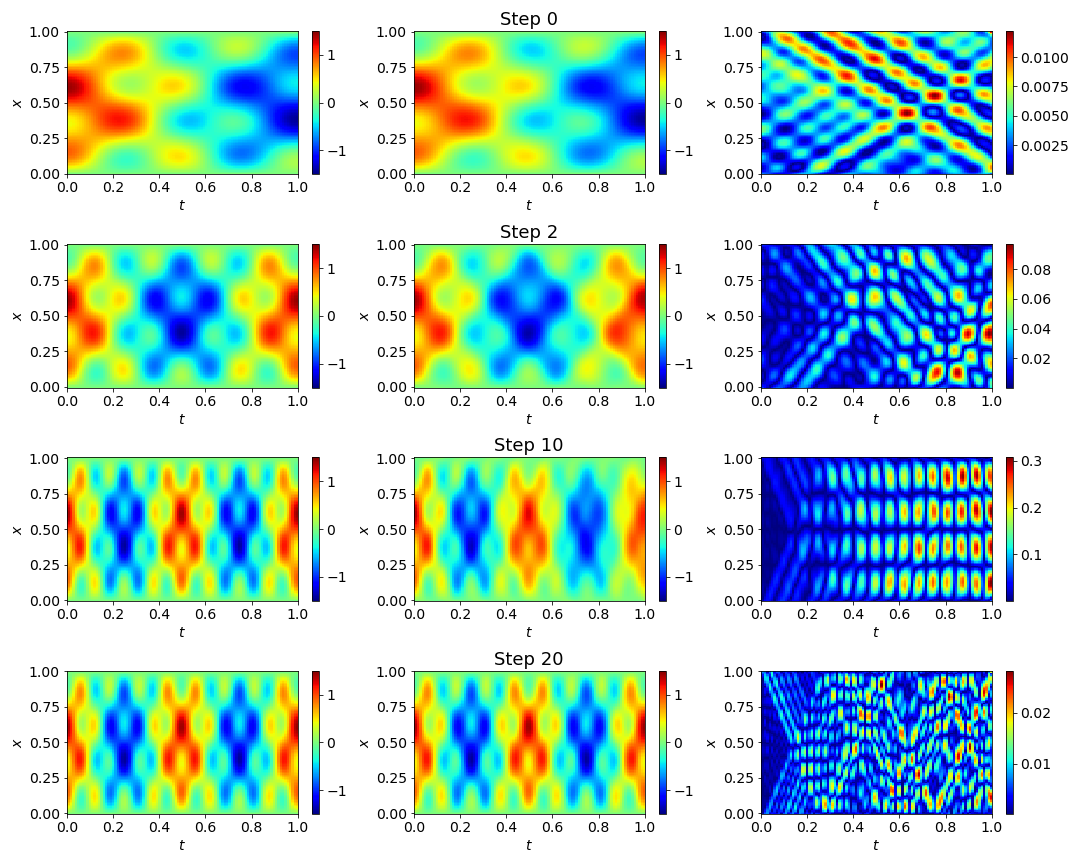}
    \caption{Results for the stacking PINN for the wave equation for Case 4. The left column has the exact solution for the value of $c$ used for a given stacking step. The results from training steps 0, 2, 10, and 15 are in the middle column. Their respective absolute errors are in the right column. }
    \label{fig:wave_c3_app}
\end{figure}

\begin{figure}[h]
    \centering
            \includegraphics[width=0.4\textwidth]{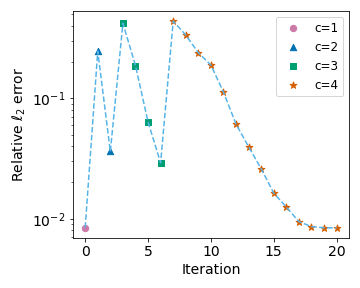}
    \caption{Relative $\ell_2$ errors for the stacking PINN for the wave equation for Case 4. The error is calculated with respect to the exact solution for the value of $c$ used for a given stacking step, which is why the relative error jumps each time $c$ is increased. }
    \label{fig:wave_c3errors_app}
\end{figure}

\clearpage
\section{Training parameters} \label{sec:appendix_trainingparams}

We report the training parameters used for all cases in this paper in Table \ref{tab:parameters_PINNs} for Sec. \ref{sec:PINN} and Table \ref{tab:parameters_DeepONet} for Sec. \ref{sec:deeponet}. 

%\begin{landscape}
\begin{table}[h]
    \centering
    \begin{tabular}{|| l | c | c |c ||} \hline
        & Sec. \ref{sec:pendulum} & Sec. \ref{sec:1st_deriv_multiscale} &Sec. \ref{sec:wave}   \\ \hline

\multicolumn{4}{l}{\textbf{PINN parameters}}   \\ \hline
Learning rate &       ($10^{-3}$, 2000, .99)    & ($10^{-3}$, 2000, .99)    & ($10^{-4}$, 2000, .99) \\
Network size &        [1, 200, 200, 200, 2]     & Varies                    & [2, 100, 100, 100, 100, 100, 1]\\
Activation function &       \texttt{swish}      & \texttt{swish}            & \texttt{tanh} \\
BC batch size &                 1               & 1                         & 300\\
Residual batch size &           200             & 400                       & 300           \\
Iterations &                    400000          & 400000                    & 400000           \\
$\lambda_{r}$ &             1.0                         & 10.0              & 1.0           \\
$\lambda_{bc}$ &            --                          & --                & 1.0           \\
$\lambda_{ic}$ &            20.0                        & 1.0               & 20.0           \\ \hline

\multicolumn{4}{l}{\textbf{Stacking parameters}}  \\ \hline
Step 0 learning rate &         ($10^{-3}$, 2000, .99)      & ($10^{-3}$, 2000, .99)    & ($10^{-4}$, 2000, .99)              \\
Step 0 network size &           [1, 100, 100, 100, 2]      & [1, 32, 32, 32, 1]        & [2, 100, 100, 100, 100, 100, 1]           \\
Nonlinear network size &   [3, 50, 50, 50, 50, 50, 2]   & [2, 16, 16, 16, 16, 1]     & [3, 100, 100, 100, 100, 100, 1]           \\
Linear network size &       [2, 20, 2]                  & [1, 5, 1]                  & [1, 1]           \\
MF learning rate &         ($10^{-3}$, 2000, .99)          & ($10^{-3}$, 2000, .99)     & ($5\times 10^{-4}$, 2000, .99)            \\
Activation function &       \texttt{swish}   &       \texttt{swish}             & \texttt{tanh}           \\
BC batch size &                 1                   & 1                         & 300           \\
Residual batch size &           200                 & 400                       & 300           \\
Iterations &                    100000              & 200000                    & 10000           \\
$\lambda_{r}$ &             1.0                         & 10.0                  & 1.0           \\
$\lambda_{bc}$ &            --                          & --                    & 1.0           \\
$\lambda_{ic}$ &            1.0                         & 1.0                   & 20.0           \\ \hline

    \end{tabular}
    \caption{Training parameters for the results in Sec. \ref{sec:PINN}. For the learning rate, the triplet $(a, b, c)$ denotes the \texttt{exponential\_decay} function in Jax \cite{jax2018github} with learning rate $a$, decay steps $b$, and decay rate $c$. The PINN parameters refer to the parameters for cases without stacking. }
    \label{tab:parameters_PINNs}
\end{table}
%\end{landscape}

%\begin{landscape}
\begin{table}
    \centering
    \begin{tabular}{|| l | c  ||} 
\multicolumn{2}{l}{\textbf{DeepONet parameters}}    \\ \hline
Learning rate &      ($10^{-3}$, 5000, .9) \\
Branch network size & $[101, 100, 100, 100, 100, 100, 100, 100]$     \\   
Trunk network size & $[2, 100, 100, 100, 100, 100, 100, 100]$        \\
Activation function &     \texttt{tanh} \\
BC batch size &         10000     \\    
Residual batch size &   10000    \\  
Iterations &            200000   \\                
$\lambda_{r}$ &         1.0 \\
$\lambda_{bc}$ &        10.0 \\
$\lambda_{ic}$ &        10.0 \\
$N_u$ &                 1000  \\ \hline
\multicolumn{2}{l}{\textbf{Stacking DeepONet parameters}}    \\ \hline
Step 0 learning rate &   ($10^{-3}$, 5000, .9)    \\
Step 0 branch network size & $[101, 100, 100, 100, 100, 100, 100, 100]$     \\   
Step 0 trunk network size & $[2, 100, 100, 100, 100, 100, 100, 100]$        \\
Step 0 iterations & 200000 for fixed $\nu$, 100000 for changing $\nu$ \\
Learning rate &    ($5\times 10^{-4}$, 5000, .95)  \\
Nonlinear branch network size & $[102, 100, 100, 100, 100, 100, 100, 100]$     \\   
Nonlinear trunk network size & $[2, 100, 100, 100, 100, 100, 100, 100]$        \\ 
Linear branch network size & $[1, 20]$     \\   
Linear trunk network size & $[2, 20]$        \\

Activation function &     \texttt{tanh} \\
BC batch size &         10000     \\    
Residual batch size &   10000    \\  
Iterations &            100000 with fixed weights, 200000 with NTK weights  \\                
$\lambda_{r}$ &         1.0 \\
$\lambda_{bc}$ &        10.0 \\
$\lambda_{ic}$ &        10.0 \\
$N_u$ &                 1000  \\ \hline
    \end{tabular}
    \caption{Training parameters for the DeepONet results in Sec. \ref{sec:deeponet}. For the learning rate, the triplet $(a, b, c)$ denotes the \texttt{exponential\_decay} function in Jax with learning rate $a$, decay steps $b$, and decay rate $c$. $N_u$ is the number of initial conditions used in the training set.}
    \label{tab:parameters_DeepONet}
    \end{table}
%\end{landscape}

\end{document}